%% file: main.tex
\documentclass{article}

\usepackage[margin=1in]{geometry}

\usepackage{microtype}
\usepackage{graphicx}
\usepackage{subfigure}
\usepackage[subfigure]{tocloft}
\usepackage{booktabs} %
\usepackage{amsmath}
\usepackage{amssymb}
\usepackage{amsfonts}
\usepackage{multirow}
\usepackage{verbatim}
\usepackage{caption}
\usepackage{longtable}
\usepackage{float}

\usepackage{enumitem}
\usepackage{tablefootnote}
\usepackage[round,semicolon]{natbib}
\usepackage{xcolor}
\usepackage{xspace}
\usepackage{textcomp}
\usepackage{makecell}
\usepackage{multirow}
\usepackage{lscape} 
\usepackage{siunitx}

\usepackage{amssymb}%

\usepackage{scrextend}

\newcommand{\modelname}[0]{{AlexaTM 20B}\xspace}

\usepackage{hyperref}

\usepackage{cleveref}
\crefname{section}{Section}{Sections}
\crefname{subsection}{Section}{Sections}
\crefname{table}{Table}{Tables}
\crefname{figure}{Figure}{Figures}
\crefname{algorithm}{Algorithm}{}
\crefname{equation}{eq.}{}
\crefname{appendix}{Appendix}{}
\usepackage{multicol}

\title{\vspace{-2em}%
  \hrule height 4pt%
  \vskip 0.25in%
  \vskip -\parskip%
  \textbf{
AlexaTM 20B: Few-Shot Learning Using a Large-Scale Multilingual Seq2seq Model}
  \vskip 0.2in%
  \vskip -\parskip%
  \hrule height 1pt%
  \vskip 0.09in}

\input{authorlist}

\date{}

\begin{document}
\maketitle

\begin{abstract}
\input{abstract}
\end{abstract}

%\newpage
%\setlength{\cftbeforesecskip}{5pt}
%\setcounter{tocdepth}{2} %
%\tableofcontents

\section{Introduction}
\label{sec:introduction}
\input{intro}

\section{Model Architecture}
\label{sec:model-architecture}
\input{model-arch}

\section{Training Data Preparation}
\label{sec:training-dataset}
\input{training-dataset}

% \section{Training Infrastructure}
% \label{sec:training-infra}
% \input{training-infra}

\section{Training Setup}
\label{sec:training-setup}
\input{training-setup}

\section{Evaluation Setups}
\label{sec:eval}
\input{evaluation}
\section{Evaluation Results}
\label{sec:results}
\input{results}

%\label{}
%\input{}

% \subsection{Reasoning}
% \label{sec:reasoning-tasks}
%\input{}

\section{Memorization}
\label{sec:memorization_analysis}
\input{memorization}

\section{Fairness and Bias Analysis}
\label{sec:fairness_analysis}
\input{fairness}

\section{Environmental Impact}
\label{sec:environmental-impact}
\input{environmental-impact}

\section{Related Work}
\label{sec:related-work}
\input{related-work}

\section{Conclusion}
\label{sec:conclusion}
\input{conclusion}

\pagebreak
\section{Acknowledgments}
\input{acknowledgements}

\bibliographystyle{icml2022}
\bibliography{main}
\clearpage
\input{appendix}

\end{document}

%% file: authorlist.tex
\author{
\textbf{Saleh Soltan\thanks{%Author contributions and ordering details are listed in \cref{sec:contributions}.\newline
Corresponding author: \texttt{ssoltan@amazon.com}.}} \quad
\textbf{Shankar Ananthakrishnan} \quad 
\textbf{Jack FitzGerald} \quad 
\textbf{Rahul Gupta} \quad
\\
\textbf{Wael Hamza} \quad 
\textbf{Haidar Khan} \quad 
\textbf{Charith Peris} \quad
\textbf{Stephen Rawls} \quad
\\
\textbf{Andy Rosenbaum} \quad 
\textbf{Anna Rumshisky} \quad 
\textbf{Chandana Satya Prakash} \quad
\\
\textbf{Mukund Sridhar} \quad 
\textbf{Fabian Triefenbach} \quad
\textbf{Apurv Verma} \quad  
\textbf{Gokhan Tur} \quad
\\
\textbf{Prem Natarajan} \quad
\vspace{0.1in}
 \\ Amazon Alexa AI }

%% file: abstract.tex
In this work, we demonstrate that multilingual large-scale sequence-to-sequence (seq2seq) models, pre-trained on a mixture of denoising and Causal Language Modeling (CLM) tasks, are more efficient few-shot learners than decoder-only models on various tasks. In particular, we train a 20 billion parameter multilingual seq2seq model called Alexa Teacher Model (\modelname) and show that it achieves state-of-the-art (SOTA) performance on 1-shot summarization tasks, outperforming a much larger 540B PaLM decoder model. \modelname also achieves SOTA in 1-shot machine translation, especially for low-resource languages, across almost all language pairs supported by the model (Arabic, English, French, German, Hindi, Italian, Japanese, Marathi, Portuguese, Spanish, Tamil, and Telugu) on Flores-101 dataset. We also show in zero-shot setting, \modelname outperforms GPT3 (175B) on SuperGLUE and SQuADv2 datasets and provides SOTA performance  on multilingual tasks such as XNLI, XCOPA, Paws-X, and XWinograd. Overall, our results present a compelling case for seq2seq models as a powerful alternative to decoder-only models for Large-scale Language Model (LLM) training.

%% file: intro.tex
Recent studies~\citep{Brown2020LanguageMA,Chowdhery2022PaLMSL} have demonstrated that Large-scale Language Models (LLMs) are capable of learning from few examples without any gradient updates (a.k.a.\ in-context learning). This capability of LLMs has been demonstrated to improve as the size of the model and the pre-training data increases~\citep{rae2021scaling,Hoffmann2022TrainingCL}. However, most  previous work has focused on decoder-only models as the architecture of choice for LLMs. The main shortcoming of decoder-only architecture is the unidirectional attention to the context~\citep{Radford2018ImprovingLU} which can be critical for long-context tasks such as summarization. Conceptually, sequence-to-sequence (seq2seq) architecture provides a better fit for training generative LLMs without losing the bidirectionally of the attention layers.  

Inspired by these considerations, in this work, we demonstrate that transformer-based seq2seq models can be used as few-shot learners, outperforming much larger decoder-only counterparts on the relevant core tasks. We pre-train a multilingual 20 billion parameter seq2seq model, which we will refer to as Alexa Teacher Model (\modelname), on a mix of denoising and Causal Language Modeling (CLM) tasks in Arabic, English, French, German, Hindi, Italian, Japanese, Marathi, Portuguese, Spanish, Tamil, and Telugu, using the Wikipedia and mC4 datasets~\citep{Xue2021mT5AM}. We follow \cite{Hoffmann2022TrainingCL} and pre-train the model for roughly 1 Trillion tokens (longer than the 300B token updates of GPT-3). \modelname is the first multilingual seq2seq model of this size with previous models having up to 11 billion parameters~\citep{Xue2021mT5AM}. 
%, among both  and decoder only models~\citep{Lin2021FewshotLW}.\footnote{While we were finalizing this paper, Huggingface~\citep{wolf-etal-2020-transformers} released a multilingual 175B parameter decoder-only model called Bloom which is yet to be fully evaluated by its authors.}
\modelname is also the first multilingual seq2seq model capable of in-context learning, since almost all previously pre-trained seq2seq models were trained exclusively on denoising tasks \citep{Raffel2020ExploringTL,Lewis2020BARTDS,Xue2021mT5AM}, and thus are not suitable for in-context learning. 

We demonstrate that not only AlexaTM 20B is capable of few-shot learning, but also outperforms much larger decoder-only models (e.g., PaLM 540B) on tasks that require attention to a long context, such as summarization. %\modelname also provides state-of-the-art (SOTA) results across almost all supported language pairs in Machine Translation (MT) in the one-shot setting. 
We show that \modelname performs better than or on par with PaLM 540B on XSUM~\citep{Narayan2018DontGM} and MLSum (de and es)~\citep{Scialom2020MLSUMTM} both in one-shot in-context learning and when fine-tuned. We also evaluate the \modelname on the SuperGLUE and SQuADv2 datasets in zero-shot setting and show that it outperforms GPT3 175B on short context tasks, despite having 8x fewer parameters.

%In Machine Translation (MT), we show that 
\modelname achieves SOTA in MT across almost all language pairs supported by the model on the Flores-101 dataset~\citep{Goyal2022TheFE}, outperforming existing supervised models using only one-shot. The gain on translation to/from low-resource languages like Marathi, Tamil, and Telugu is significant (e.g., 21.8 Arabic to Tamil spBleu score compared to 0.9 score of the supervised M2M-124 615M model \citep{Goyal2022TheFE}). The results  suggest that large-scale seq2seq-style pre-training, as formulated in this work, may be the best way to improve low-resource language MT with limited training pairs, especially when a large amount of monolingual data is available. \modelname also performs very well translating directly from different languages, in contrast to many-to-many MT systems that are trained mainly on to/from English language pairs \citep{Fan2021BeyondEM}. We believe these results suggest a paradigm shift in how MT systems should be developed.

Additionally, we evaluate \modelname on multilingual tasks including XNLI~\citep{Conneau2018XNLIEC}, XCOPA~\citep{Ponti2020XCOPAAM}, Paws-X~\citep{Yang2019PAWSXAC}, and XWinograd~\citep{Tikhonov2021ItsAI}. We show that \modelname provides SOTA performance in zero-shot setting for all of these tasks, across all supported languages, improving on previous SOTA achieved by the XGLM 7.5B model \citep{Lin2021FewshotLW}.

In order to understand the risks associated with using this model, we also analyze the amount of memorization in \modelname and assess its fairness and biases. In contrast to \citet{Carlini2022QuantifyingMA}’s observations for GPT3, \modelname's memorization slightly decreases with increasing context size in both the CLM and denoising modes. This suggests that the denoising objective (perhaps, coupled with multilinguality of the model) may be breaking the memorization of longer contexts by the model.   
%Nevertheless, this can be counted as a positive feature of the model.
While the ability to memorize training data can in principle degrade model performance on some tasks, it also alleviates some of the privacy concerns associated with training LLMs. 

With respect to bias and toxicity, the model provides SOTA results on Winogender bias benchmark in zero-shot setting. However, consistent with prior work, we observe that in open-ended generation, \modelname Toxicity Probability of Continuation (TPC) increases with Toxicity Probability of Prompt (TPP) (i.e.,\ toxic prompts lead the model to generate more toxic continuations). Although, TPC is lower than TPP for the majority of the cases and that TPC is lower than the human baseline for low toxicity prompts. %However, as prompt toxicity increases, TPC surpasses human baseline indicating that model starts generating really toxic continuations with increasing prompt toxicity.

Overall, we demonstrate that the proposed style of pre-training of seq2seq models shows better performance compared to much larger decoder-only LLMs across different tasks, both in a few-shot setting and with fine-tuning. Additionally, we did not observe any negative results relative to decoder-only LLMs of similar size. We hope our work presents a compelling case for seq2seq models as a powerful alternative to decoder-only models for LLM training.

To summarize, the main contributions of our work are four-fold: 1) we train and release the largest available multilingual seq2seq model on a mix of denoising and CLM tasks and show that it is capable of few-shot in-context learning, 2) we demonstrate that large-scale seq2seq models are better at in-context learning when the context is long (e.g., summarization) compared to much larger decoder-only models, 3) we demonstrate that an 8x smaller seq2seq model can peform on par or better than GPT3 175B on SuperGLUE and SQuAD benchmarks, suggesting the efficiency of seq2seq models training, and 4) we show the strength of seq2seq multilingual pre-training in one-shot MT, suggesting a new paradigm for  non-English-centric MT and low-resource languages.

We will release the \modelname model on \url{https://github.com/amazon-research/alexa-teacher-models}.

%% file: model-arch.tex
For \modelname, we used the standard Transformer model architecture \citep{Vaswani2017AttentionIA} with learnable positional embeddings with the small modification of moving the layernorms (both in the encoder and the decoder) to be located exactly at the beginning of each layer (right after the skip connection) instead of at the end of each layer (i.e., Pre-LN). This modification has been demonstrated to improve the stability of the training, especially for large models~\citep{Shoeybi2019MegatronLMTM,Xiong2020OnLN}. Table~\ref{tab:model-hyperparams} shows the detailed hyper-parameters of the \modelname's architecture.

\begin{table}[h!]
\begin{center}
\begin{tabular}{lccccc} 
\toprule
Model & Encoder Layers & Decoder Layers  & \# of Heads   &  $d_\textrm{model}$ &  \makecell[c]{\# of Parameters \\(in billions)}  \\ 
\midrule
\modelname & 46 & 32 & 32 & 4096 & 19.75 \\
\bottomrule
\end{tabular}
\end{center}
\caption{Model architecture details. The feed-forward size $d_\textrm{ff}$ is $4 \times d_\textrm{model}$.}
\label{tab:model-hyperparams}
\end{table}

%% file: training-dataset.tex
\subsection{Datasets}

The pre-training data consists of Wikipedia and mC4~\citep{Xue2021mT5AM} datasets. We use the data in 12 languages, namely, Arabic, English, French, German, Hindi, Italian, Japanese, Marathi, Portuguese, Spanish, Tamil, and Telugu. We pack sequences of tokens to produce sequences of approximately 1024 subword units. We allow unrelated content to be packed together in the same sequence but we separate them with a special symbol ({\ttfamily [DOC]}). Maintaining a relatively constant number of subword sequences reduces padding and results in efficient compute. 

\begin{table}[h!]
\begin{center}
\begin{tabular}{lccccc} 
\toprule
%Model & Encoder Layers & Decoder Layers  & \# of Heads   &  $d_\textrm{model}$ &  %\makecell[c]{\# of Parameters \\(in billions)}  \\ 
Dataset & Sampling & Final Tokens & Train Set Percent \\
\midrule
%\modelname & 46 & 32 & 32 & 4096 & 19.75 \\
Wikipedia & \makecell[l]{$\alpha=0.5$ lang sampling, then \\ 7 $\times$ spoken form plus \\ 3 $\times$ written form} & 119B & 9\% \\
mC4 & \makecell[l]{$\alpha=0.5$ lang sampling, then \\ 0.7 $\times$ spoken form plus \\ 0.3 $\times$ written form} & 1.2T & 91\% \\
\bottomrule
\end{tabular}
\end{center}
\caption{Source datasets and sampling methods for the training set.}
\label{tab:pretrain-dataset}
\end{table}

Since our objective is to allow \modelname to act on both spoken queries and written text, we use a written to spoken formatter for all languages to format the data into spoken format (remove capitalization, remove punctuations, etc.). We include more spoken than written text to satisfy our internal use cases. The final training set, described in Table \ref{tab:pretrain-dataset}, is developed by combining data from various sources using three types of mixing:
\begin{itemize}
    \item Frequency based upsampling which helps increase the representation of under-represented languages following~\cite{Conneau2020UnsupervisedCR}. In particular, we sample sentences according to a multinomial distribution with probabilities $(q_1, q_2, \dots, q_N)$, where:
\begin{align}
q_i=\frac{p_i^\alpha}{\sum_{j=1}^N p_j^\alpha}, ~p_i=\frac{n_i}{\sum_{j=1}^N n_j}
\end{align}
in which $N$ is the total number of languages and $n_i$ is the total number of sentences in language $i$ (we set $\alpha=0.5$).
    \item Upsampling Wikipedia data (which has a higher quality) by 10 to be represented more in all data.
    \item Scaling to favor spoken format over written 7 to 3.
\end{itemize}

We followed~\cite{Brown2020LanguageMA} and filtered out known benchmarks from the training data by checking for 13-gram overlaps between each sentence and the benchmarks (for the list of filtered datasets see Appendix~\ref{sec:filtered}).

\subsection{Subword Tokenizer}

We use SentencePiece (SP)~\citep{Kudo2018SentencePieceAS} to tokenize the input. We trained a unigram-based model of SP to determine a set of sub-words that best represents the training data. Our final choice was to train a 150K unigram sentencepiece model from a 7 to 3 mixture ratio of spoken to written data. We reserve 1K vocabulary entries to be used for tags in downstream tasks (e.g., parse nodes in semantic parsing task). We also manually augment the vocabulary with a few entries to guarantee better character and word coverage.

%% file: training-setup.tex
\modelname model class is derived from BART~\citep{Lewis2020BARTDS} class implementation in Huggingface~\citep{Wolf2019HuggingFacesTS} allowing us to benefit from the {\ttfamily generate} function built in the parent class for inference. To train the model, we used a denoising objective in which we drop 15\% of the tokens in the input (in spans of length determined  by a Poisson distribution with mean equal to 3) and expect the model to reconstruct the input. We do not add any mask tokens in the input during training: 1) to have the most consistency during pre-training, inference, and fine-tuning (i.e., no mask tokens appear in any setting), 2) to require the decoder to play a more active role during pre-training, and 3) to leverage the 10B encoder that we had trained previously~\citep{FitzGerald2022AlexaTM} to initialize \modelname's encoder (adding [MASK] would have made decoder job easy given encoder's ability in ``unmasking").

\begin{figure}
    \centering
    \includegraphics[scale=0.5]{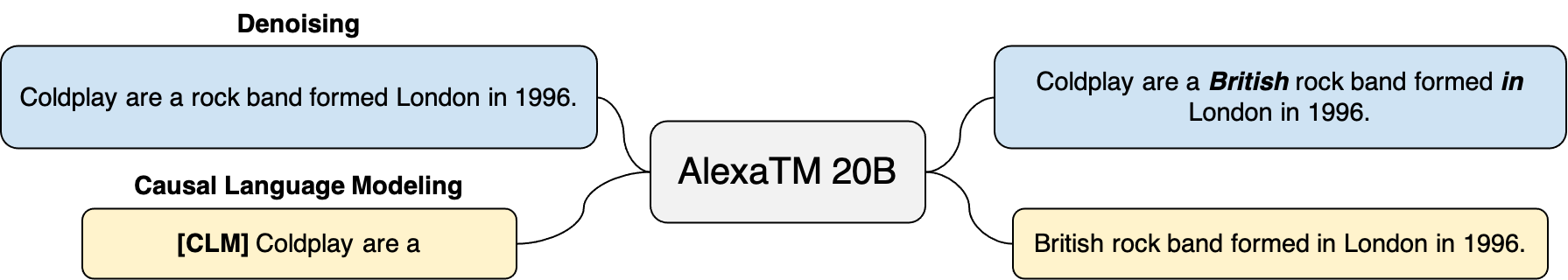}
    \caption{\modelname pre-training objectives. During pre-training the model is trained on the denoising task 80\% and on the Causal Language Modeling 20\% of the time. pre-training data consists of Wikipedia and mC4 datasets in 12 languages as specified in Section~\ref{sec:training-dataset}.}
    \label{fig:pre-training}
\end{figure}

To make our model more efficient for in-context learning, we added an extra Causal Language Modeling (CLM) task to the 20B training for 20\% of the time.\footnote{The CLM using seq2seq models has also been called Prefix Language Modeling (PLM)~\citep{Raffel2020ExploringTL}. However, since the term ``prefix'' has been excessively used recently in other context~\citep{Li2021PrefixTuningOC}, we prefer to use CLM to avoid any confusions.} In this task, the model is required to continue the input instead of denoising the input. The model will know to do CLM based on a special token that we add to the beginning of the sentence ({\ttfamily [CLM]}). For the CLM task, we only feed a single document (instead of concatenation of multiple documents) and give 20\% to 80\% of the document randomly (uniformly) as the input to the model (so the model learns to continue both from long inputs and short inputs).
%Although we have an extra task when we train the 20B, we only monitor the denoising task for validation since that is the main objective during pre-training.

%2.3. Initialize from AlexaTM pre-trained Encoders

To speed up the model training, we initialized the encoder by an internal 10B pre-trained encoder~\cite{FitzGerald2022AlexaTM} (we also initialize the decoder embeddings and LM head embeddings with the embedding from encoder but we do not tie any pair of embeddings). During training, we initially keep the encoder frozen (for around 100k updates) but unfreeze the encoder to train the model end to end. %once the cross-entropy loss is stabilized (i.e., move slowly).

We trained \modelname for 120 days on 128 A100 GPUs for the total of 500k updates with the accumulated batch size of 2 million tokens (total of 1 trillion token updates). We used Adam optimizer~\citep{Kingma2015AdamAM} with $lr=1e^{-4}$ with linear decay to $lr=5e^{-6}$ over 500k updates. We used weight decay of $0.1$ on all parameters except biases and layernorms. Finally, we trained the model in {\ttfamily BFloat16} which helped with the stability of training~\citep{Raffel2020ExploringTL}.

We used DeepSpeed's ZeRO Stage 3~\citep{Rasley2020DeepSpeed} to partition model weights, optimizer states, and gradients across all GPU workers, allowing us to train the model with high throughput. We relied on an internal and optimized version of DeepSpeed that we have since open-sourced~\citep{chiu_zheng_2022} to obtain training throughput of up to 154 TFLOPS/GPU on 16 AWS p4d.24xlarge compute instances.

%% file: evaluation.tex
We evaluate \modelname both using zero/few-shot in-context learning as well as by finetuning the model on selected generation tasks. In all few-shot learning settings, we use greedy search. 
\subsection{Few-shot Learning}
Since \modelname is trained both on denoising and CLM tasks, both of these modes can be used for in-context learning. In this subsection, we describe some of the techniques that we used for in-context learning using the model.

For some tasks, instead of asking the model to generate answers, we present multiple inputs to the model (corresponding to task labels/choices) and calculate the model's score for each input. We refer to this as scoring. Specifically, we provide the encoder and decoder inputs to \modelname (e.g. see Table~\ref{tab:prompt_examples2}) and compute the cross-entropy loss in the decoder with teacher-forcing for each set of inputs. The model predicts the task label or choice with the best score.

\subsubsection{Denoising Mode}
The denoising mode can be used both for generation (by dropping a few tokens from end of a prompt) and scoring (by dropping tokens from the input and comparing the loss for different output choices). Notice that since during pre-training we dropped only short spans, the generation in denoising mode is most effective when only a few tokens are omitted in the input.
\subsubsection{CLM Mode}
The CLM mode is similar to the way previous large-scale decoder-only models have been using these models for generation and scoring for in-context learning~\citep{Brown2020LanguageMA,Chowdhery2022PaLMSL,Lin2021FewshotLW}. We follow these works and use the CLM mode of \modelname for various tasks. The CLM mode can also be used for scoring when the task provides choices to select from. 
\subsubsection{CLM Mode with Fusion in the Decoder}
One benefit of seq2seq models compared to decoder-only models is that we can fit more shots into their context using Fusion-in-Decoder (FiD) idea which was initially proposed by \citet{Izacard2021LeveragingPR}. For this and in few-shot settings, we encode many 1-shot examples using the encoder and let the decoder attend to all of these examples while generating the output. Throughout this paper, when we use FiD idea, we denote number of shots by $^{*}$ such as $32^*$-shots. 
\subsection{Fine-tuning}
We also finetune \modelname on a selected generation tasks to compare its performance to previous seq2seq as well as much larger decoder-only models (when finetuned). For all the finetunings, we add {\ttfamily [CLM]} to the beginning of the input (basically finetuning the model in the CLM mode, although denoising mode could also be used but may require extra number of training steps). For all the tasks that we finetune \modelname, we use adam with $lr=1e^{-6}$ with linear decay to $lr=1e^{-7}$ and picked the best checkpoint based on validation perplexity. Given the finetuning cost, we did not do any task specific hyper-parameter searches.

%% file: results.tex
In this section, we evaluate the \modelname performance across monolingual and multilingual tasks. In particular, we demonstrate that \modelname performs better or in par to the largest dense decoder-only model to date (i.e., PaLM 540B) in summarization both in 1-shot and fine-tuning settings. We also show that \modelname provides the best scores in Flores 101 Machine Translation (MT) in almost all languages pairs supported by the model (especially for low-resource languages like Marathi, Tamil, and Telugu). Moreover, it performs on par or better than GPT3 175B parameter model across various English tasks in zero-shot (e.g., SuperGLUE). Nevertheless, we demonstrate  that there are different tasks like ``reasoning'' that the model cannot do as well as much larger models.

Additionally, we evaluate \modelname on a few multilingual datasets including XNLI, XCOPA, Paws-X, and XWinograd and show that \modelname achieves SOTA numbers in zero-shot on all these tasks across all languages better than XGLM 7.5B model~\citep{Lin2021FewshotLW}. The prompts used for each task and the evaluation mode are provided in Appendix~\ref{sec:prompt_examples}.

\subsection{Multilingual Natural Language Generation}
\label{sec:nlg}

\input{nlg}

\subsection{Machine Translation}
\label{sec:translation-tasks}
\input{mt}

\clearpage

\subsection{Multilingual NLP Tasks}
\label{sec:cross-lingual}
\input{cross_lingual}

\subsection{English NLP Tasks}
\label{sec:english_nlp}
\input{english_nlp}

\subsection{Shortcoming on Reasoning Tasks}
\label{sec:shortcoming}
\input{shortcomings}

%% file: nlg.tex
\begin{table}[h]
    \setlength{\tabcolsep}{6pt}
    \centering
    \small
    \begin{tabular}{p{2cm}cccccccccc}
    \toprule
    & \multicolumn{4}{c}{1-shot} & \multicolumn{6}{c}{Finetuning} \\
    \cmidrule(l{3pt}r{3pt}){2-5} \cmidrule(l{3pt}r{3pt}){6-11}
    Task  & \makecell[c]{PaLM \\8B} & \makecell[c]{PaLM \\62B} & \makecell[c]{PaLM \\540B} & \makecell[c]{AlexaTM \\20B}  & \makecell[c]{T5 \\XXL} & \makecell[c]{PaLM \\8B} & \makecell[c]{PaLM \\62B} &  \makecell[c]{PaLM \\540B} & \makecell[c]{AlexaTM \\20B}& SOTA\\
    \midrule
    MLSum (de)  & $4.6$ & $10.5$ & $12.8$ & $\textbf{28.51}$ & \underline{${35.9}$} & $26.5$ & $30.0$ & $33.1$ & \underline{$33.73$} & $\textbf{36.4}^a$\\
    MLSum (es)  & $2.3$ & $3.2$ & $3.6$ & $\textbf{5.22}$  & $12.0$ & $10.6$ & $11.2$ & $12.0$ &\underline{13.59} & $\textbf{13.8}^a$\\
    MLSum (fr)  & - & - & - & $\textbf{10.24}$  & - & - & - & - &\textbf{26.09} & $10.61^b$\\
    XSum (en)  & $7.9$ & $11.2$ & $\textbf{12.2}$ &$11.69$ & $21.0$ & $16.3$ & $18.5$ & $21.2$ &\underline{24.16} &$\textbf{27.1}^c$\\
    \bottomrule
    \end{tabular}
    \caption{ROUGE-2 results on summarization datasets. T5 and PaLM scores are from~\cite{Chowdhery2022PaLMSL}. Bold numbers denote the best and underline denotes better than PaLM 540B. The SOTA numbers are from $^a$\citep{Gehrmann2021TheGB}, $^b$\citep{Scialom2020MLSUMTM}, and $^c$\citep{Zoph2022STMoEDS}.} 
    
    \label{tab:sum-table}
\end{table}

\begin{table}[h]
    \setlength{\tabcolsep}{6pt}
    \centering
    \small
    \begin{tabular}{p{2cm}cc}
    \toprule
    & \multicolumn{2}{c}{1-shot}  \\
    \cmidrule(l{3pt}r{3pt}){2-3} 
    Task  &  \makecell[c]{PaLM \\540B} & \makecell[c]{AlexaTM \\20B}  \\
    \midrule
    MLSum (de)  & 18.0 & \textbf{34.21} \\
    MLSum (es)  & 10.2 &  \textbf{14.63} \\
    XSum (en)  & 24.4 & \textbf{25.73} \\
    \bottomrule
    \end{tabular}
    \caption{ROUGE-L results on summarization datasets comparing PalM 540B and AlexaTM 20B. PaLM scores are only provided for the 540B model by~\cite{Chowdhery2022PaLMSL}.} 
    \label{tab:sum-table-L}
\end{table}

In this section, we evaluate \modelname model performance on generation tasks. For comparison we focus on the tasks that the PaLM models~\citep{Chowdhery2022PaLMSL} were evaluated on.  Table~\ref{tab:sum-table} presents the ROUGE-2 scores on different summarization tasks both in 1-shot as well as when finetuned.\footnote{All the Rouge scores are computed using {\ttfamily rouge-score 0.0.4} python package~\citep{Lin2004ROUGEAP} with {\ttfamily use\_stemmer} option set to {\ttfamily true} following~\cite{Gehrmann2021TheGB}.} As can be seen, \modelname outperforms PaLM 540B model across all tasks when finetuned, indicating that for tasks that require attention to a long context, even a 25 times larger decoder-only model cannot compensate for a lack of bidirection attention mechanism.

Moreover, although one would have expected that a finetuned seq2seq model should perform better than a finetuned decoder-only (rather much larger) model, as can be seen in the Tables~\ref{tab:sum-table} and \ref{tab:sum-table-L}, even in the 1-shot setting the \modelname performs better than PaLM 540B (except only in ROUGE-2 on XSum). These results reveal an inherent weakness of decoder-only models on tasks that have a long context.

\begin{table}[h]
    \setlength{\tabcolsep}{6pt}
    \centering
    \small
    \begin{tabular}{p{2cm}ccccccccc}
    \toprule
    & \multicolumn{3}{c}{1-shot} &\multicolumn{1}{c}{Few-shot}& \multicolumn{5}{c}{Finetuning} \\
    \cmidrule(l{3pt}r{3pt}){2-4} \cmidrule(l{3pt}r{3pt}){5-5} \cmidrule(l{3pt}r{3pt}){6-10}
    Task  & \makecell[c]{PaLM \\62B} & \makecell[c]{PaLM \\540B} & \makecell[c]{AlexaTM \\20B}  & \makecell[c]{AlexaTM \\20B}  & \makecell[c]{T5 \\XXL}  & \makecell[c]{PaLM \\62B} &  \makecell[c]{PaLM \\540B} & \makecell[c]{AlexaTM \\20B} & SOTA\\
    \midrule
    E2E (en)  & $33.5$ & $\textbf{35.2}$ & $21.25$ & \underline{$34.25$} & $45.3$ & $45.2$ & $45.3$ & \underline{$45.3$} & $\textbf{45.8}^a$\\
    WebNLG (en) & \underline{$38.6$} & \textbf{$44.4$} & $20.0$  & $26.16$& $39.6$ & $48.6$ & \textbf{$49.3$} & \underline{$49.0$} & $\textbf{53.5}^b$\\
    \bottomrule
    \end{tabular}
    \caption{ROUGE-2 results on NLG datasets. Underline denotes the best between models with less than 62B parameters. Bold denotes the best score. The few-shot results for \modelname is based on $32^{*}$ shots. The SOTA numbers are from $^a$\citep{Xue2021mT5AM} and $^b$\citep{Bakshi2021StructuretoTextGW}.} 
    
    \label{tab:nlg-table}

\end{table}

We also evaluated \modelname on Cleaned E2E NLG~\citep{Novikova2017TheED,Dusek2019SemanticNM} and WebNLG 2020~\citep{Ferreira2020The2B} datasets which has much shorter context than the summarization task. As can be seen in Table~\ref{tab:nlg-table}, for these tasks, the decoder-only models perform much better than \modelname (although we can bridge the gap by providing the model with more shots). Nevertheless, as in the summarization case, in the finetuned case, the \modelname performs on par or better than PaLM 62B and 540B models.

\begin{table}[h]
    \setlength{\tabcolsep}{6pt}
    \centering
    \small
    \scalebox{0.92}{
    \begin{tabular}{p{2cm}ccccc}
    \toprule
    & \multicolumn{3}{c}{1-shot (R1/R2/RL)} & \multicolumn{2}{c}{5-shot (R1/R2/RL)}\\
    \cmidrule(l{3pt}r{3pt}){2-4} \cmidrule(l{3pt}r{3pt}){5-6}
    Task  &  \makecell[c]{AlexaTM 20B}  & \makecell[c]{T5 XXL \\ LM-adapt} & \makecell[c]{UL2 20B} & \makecell[c]{AlexaTM 20B}  & \makecell[c]{T5 XXL \\ LM-adapt}\\
    \midrule
    MLSum (de)  & \textbf{38.11}/\textbf{28.51}/\textbf{34.21} & 28.24/18.81/24.63 & -- & -- & --\\
    MLSum (es) & \textbf{18.82}/\textbf{5.22}/\textbf{14.63} & 14.18/3.5/11.15 & -- & -- & -- \\
    MLSum (fr)  & \textbf{25.85}/\textbf{10.24}/\textbf{19.03} & 19.99/7.31/14.25 & -- & -- & -- \\
    XSum (en)  & \textbf{32.41}/\textbf{11.69}/\textbf{25.73} & 31.14/10.76/24.38 & 25.5/8.6/19.8 & &\\
    \midrule
    E2E (en)  & 45.93/21.25/32.36 & \textbf{53.25}/\textbf{24.95}/\textbf{36.68} & -- & 52.46/\textbf{28.8}/\textbf{41.15} & \textbf{53.21}/22.81/34.02\\
    WebNLG (en)  & 34.47/20/\textbf{29.22} & \textbf{37.44}/\textbf{20.78}/29.04 & -- & \textbf{45.07}/\textbf{27.61}/\textbf{34.35} & 38.49/22.39/30.13\\
    \bottomrule
    \end{tabular}
    }
    \caption{Few-shot performance comparison in ROUGE score between \modelname and T5 XXL LM-adapt~\citep{Lester2021ThePO}. We also included 1-shot results for XSum for UL2 20B from~\cite{Tay2022UnifyingLL}.} 
    \label{tab:t5-compare-table}
\end{table}

Finally, we compared \modelname to T5 XXL (LM-Adapt)~\citep{Lester2021ThePO} and a very recent UL2 20B~\citep{Tay2022UnifyingLL} that are trained very similarly to our model. In particular, T5 XXL (LM-adapt) is T5 model~\citep{Raffel2020ExploringTL} which has been train on CLM task for extra 100k updates (for 1.1 trillion token updates in total including the T5 pre-training). The UL2 20B is also a seq2seq model which has been trained (concurrently to our work) on a mix of CLM and denoising tasks on English only data for 1 trillion token updates. We report UL2 20B numbers for XSum from the paper but computed the T5 XXL (LM-adapt) ourselves. As can be seen in Table~\ref{tab:t5-compare-table}, the \modelname outperforms both T5 XXL (LM-adapt) and UL2 20B in 1-shot summarization. In contradiction to what reported by~\cite{Tay2022UnifyingLL}, however, we observe that T5 XXL (LM-adapt) performs very strongly across different tasks. Especially, it outperforms UL2 20B in 1-shot summarization on XSum and PaLM 62B on MLSUM (es and de). This may suggest why the prompt tuning idea of \cite{Lester2021ThePO} worked on T5 XXL (LM-adapt), since the model is already a good few-shot learner (and why it probably did not work as well on the original T5 model). 

The T5 XXL (LM-adapt) also performs very well on E2E and WebNLG tasks. In particular it performs better than \modelname in 1-shot settings in E2E. However, in both tasks \modelname performs better in few-shot settings. Since UL2 20B did not report on these tasks, we only compared to T5 XXL (LM-adapt). Overall, performance evaluation of our model and T5 XXL (LM-adapt) on generation tasks demonstrates that seq2seq models are as good as or better than decoder-only models that are limited by their unidirectional attention.

%% file: mt.tex
\begin{table*}[t!]
\centering
\begin{small}
\addtolength{\tabcolsep}{-2.5pt}
%\scalebox{0.9}{
\begin{tabular}{lrcccccccccccc}
\toprule
&& shots & ar & fr & en & de & it & ja & hi & mr & ta & te & es  \\
\midrule
\multirow{3}{*}{ar}
& Supervised & NA & -- & 25.7 & 25.5 & 18.7 & 17.8 & 16.0 & 19.4 & 2.5 & 0.9 & 0.3 & 16.74  \\
%& GPT-3 6.7B & 32 & & & & & & & & & & & & \\
& XGLM 7.5B & 32 & -- & 17.9 & 27.7 & 12.2  & -- & -- & 7.8 & -- & 3.7 & -- & --  \\
& \modelname & 1 & -- & \textbf{\underline{35.5}} & \textbf{\underline{41.8}} & \textbf{\underline{27.5}} & \textbf{\underline{25.4}} & \textbf{\underline{20.6}} & \textbf{\underline{24.4}} & \textbf{\underline{15.9}} & \textbf{\underline{21.8}} & \textbf{\underline{6.0}} & \textbf{\underline{23.2}}  \\
\midrule
\multirow{3}{*}{fr}
& Supervised & NA & 15.4 & -- & 37.2 & 28.5 & 28.6 & 21.5 & 22.9 & 6.9 & 0.8 & 0.6 & 25.6  \\
%& GPT-3 6.7B & 32 & & & & & & & & & & & & \\
& XGLM 7.5B & 32 & 5.9 & -- & 40.4 & 20.4 & & & 13.7 & -- & 6.6 & -- & --  \\
& \modelname & 1 & \textbf{\underline{24.7}} & -- & \textbf{\underline{47.1}} & \textbf{\underline{32.4}} & \textbf{\underline{29.9}} & \textbf{\underline{24.3}} & \textbf{\underline{27.3}} & \textbf{\underline{19.3}} & \textbf{\underline{23.7}} & \textbf{\underline{27.0}} & \textbf{\underline{26.3}}   \\
\midrule
\multirow{3}{*}{en}
& Supervised & NA & 17.9 & 42.0 & -- & 32.6 & 27.7 & 22.8 & 28.1 & 10.4 & 3.4 & 1.9 & 25.6  \\
%& GPT-3 6.7B & 32 & & & & & & & & & & & & \\
& XGLM 7.5B & 32 & 11.5 & 36.0 & -- & 27.6 & -- & -- & 19.9 & -- & 8.5 & -- & --  \\
& \modelname & 1 & \textbf{\underline{32.0}} & \textbf{\underline{50.7}} & -- & \textbf{\underline{41.2}} & \textbf{\underline{34.4}} & \textbf{\underline{28.4}} & \textbf{\underline{35.1}} & \textbf{\underline{24.7}} & \textbf{\underline{30.0}} & \textbf{\underline{34.2}} & \textbf{\underline{31.0}}  \\
\midrule
\multirow{3}{*}{de}
& Supervised & NA & 14.8 & 35.5 & 35.8 & -- & 25.9 & 21.1 & 23.4 & 9.2 & 2.3 & 0.6 & 23.4  \\
%& GPT-3 6.7B & 32 & & & & & & & & & & & & \\
& XGLM 7.5B & 32 & 6.6 & 27.9 & 35.8 & -- & -- & -- & 14.3 & -- & 4.8 & -- & --  \\
& \modelname & 1 & \textbf{\underline{24.3}} & \textbf{\underline{38.7}} & \textbf{\underline{45.5}} & -- & \textbf{\underline{29.4}} & \textbf{\underline{24.9}} & \textbf{\underline{27.6}} & \textbf{\underline{18.7}} & \textbf{\underline{24.1}} & \textbf{\underline{27.6}} & \textbf{\underline{26.1}}  \\
\midrule
\multirow{3}{*}{it}
& Supervised & NA & 13.4 & 34.4 & 28.7 & 24.2 & -- & 19.8 & 20.6 & 9.0 & 2.2 & 0.5 & 24.5  \\
%& GPT-3 6.7B & 32 & & & & & & & & & & & & \\
& XGLM 7.5B & 32 & -- & -- & -- & -- & -- & -- & -- & -- & -- & -- & --  \\
& \modelname & 1 & \textbf{\underline{22.0}} & \textbf{\underline{35.7}} & \textbf{\underline{37.5}} & \textbf{\underline{27.9}} & -- & \textbf{\underline{22.9}} & \textbf{\underline{24.7}} & \textbf{\underline{15.8}} & \textbf{\underline{21.2}} & \textbf{\underline{24.8}} & \textbf{\underline{25.9}}  \\
\midrule
\multirow{3}{*}{ja}
& Supervised & NA & 10.3 & 21.9 & 19.5 & 16.3 & 16.0 & -- & \underline{17.9} & 7.6 & \underline{3.1} & \underline{0.5} & 15.7  \\
%& GPT-3 6.7B & 32 & & & & & & & & & & & & \\
& XGLM 7.5B & 32 & -- & -- & -- & -- & -- & -- & -- & -- & -- & -- & --  \\
& \modelname & 1 & \textbf{\underline{12.6}} & \textbf{\underline{27.0}} & \textbf{\underline{28.5}} & \textbf{\underline{21.2}} & \textbf{\underline{21.3}} & -- & \textbf{16.7} & \textbf{\underline{15.5}} & 0.2 & 0.2 & \textbf{\underline{20.0}} \\
\midrule
\multirow{3}{*}{hi}
& Supervised & NA & \underline{12.2} & 25.9 & 27.9 & 19.4 & 17.9 & 18.0 & -- & 12.6 & 3.8 & 0.7 & 16.6  \\
%& GPT-3 6.7B & 32 & & & & & & & & & & & & \\
& XGLM 7.5B & 32 & 6.1 & 15.4 & 25.2 & 12.3 & -- & -- & -- & -- & 1.9 & -- & --  \\
& \modelname & 1 & \textbf{8.9} & \textbf{\underline{32.8}} & \textbf{\underline{40.0}} & \textbf{\underline{25.4}} & \textbf{\underline{23.3}} & \textbf{\underline{20.9}} & -- & \textbf{\underline{15.9}} & \textbf{\underline{24.5}} & \textbf{\underline{23.3}} & \textbf{\underline{21.5}}  \\
\midrule
\multirow{3}{*}{mr}
& Supervised & NA & 7.4 & 16.6 & 18.7 & 12.6 & 12.4 & \underline{13.2} & 21.3 & -- & 4.4 & 0.5 & 11.8  \\
%& GPT-3 6.7B & 32 & & & & & & & & & & & & \\
& XGLM 7.5B & 32 & -- & -- & -- & -- & -- & -- & -- & -- & -- & -- & --  \\
& \modelname & 1 & \textbf{\underline{14.1}} & \textbf{\underline{29.3}} & \textbf{\underline{35.7}} & \textbf{\underline{22.8}} & \textbf{\underline{21.4}} & \textbf{12.0} & \textbf{\underline{27.6}} & -- & \textbf{\underline{15.7}} & \textbf{\underline{23.5}} & \textbf{\underline{20.6}}  \\
\midrule
\multirow{3}{*}{ta}
& Supervised & NA & 1.1 & 6.8 & 8.3 & 4.9 & 5.7 & 2.4 & 6.9 & 3.1 & -- & 0.3 & 5.3  \\
%& GPT-3 6.7B & 32 & & & & & & & & & & & & \\
& XGLM 7.5B & 32 & 5.4 & 10.3 & 16.3 & 8.4 & -- & -- & 7.2 & -- & -- & -- & --  \\
& \modelname & 1 & \textbf{\underline{18.2}} & \textbf{\underline{27.6}} & \textbf{\underline{32.3}} & \textbf{\underline{21.5}} & \textbf{\underline{20.6}} & \textbf{\underline{19.3}} & \textbf{\underline{25.0}} & \textbf{\underline{18.4}} & -- & \textbf{\underline{26.9}} & \textbf{\underline{19.1}}  \\
\midrule
\multirow{3}{*}{te}
& Supervised & NA & 4.8 & 13.2 & 15.1 & 8.8 & 8.7 & 8.8 & 12.9 & 6.7 & 3.2 & -- & 9  \\
%& GPT-3 6.7B & 32 & & & & & & & & & & & & \\
& XGLM 7.5B & 32 & -- & -- & -- & -- & -- & -- & -- & -- & -- & -- & --  \\
& \modelname & 1 & \textbf{\underline{19.1}} & \textbf{\underline{26.7}} & \textbf{\underline{38.8}} & \textbf{\underline{23.8}} & \textbf{\underline{22.3}} & \textbf{\underline{14.9}} & \textbf{\underline{26.7}} & \textbf{\underline{20.7}} & \textbf{\underline{18.5}} & -- & \textbf{\underline{20.9}}  \\
\midrule
\multirow{3}{*}{es}
& Supervised & NA & 12.1 & 29.3 & 25.1 & 21.0 & 23.9 & 18.1 & 18.5 & 7.1 & 0.4 & 0.5 & --  \\
%& GPT-3 6.7B & 32 & & & & & & & & & & & & \\
& XGLM 7.5B & 32 & -- & -- & -- & -- & -- & -- & -- & -- & -- & -- & --  \\
& \modelname & 1 & \textbf{\underline{20.8}} & \textbf{\underline{33.4}} & \textbf{\underline{34.6}} & \textbf{\underline{25.8}} & \textbf{\underline{26.7}} & \textbf{\underline{22.3}} & \textbf{\underline{24.3}} & \textbf{\underline{17.8}} & \textbf{\underline{21.2}} & \textbf{\underline{23.7}} & --  \\
\bottomrule
\end{tabular}
%}
\end{small}
\caption{Machine translation results on FLORES-101 devtest (spBLEU). Source language in rows, target language in columns.  XGLM 7.5B  use 32 examples from the dev set for few-shot learning while \modelname uses only 1-shot. Supervised results correspond to the M2M-124 615M model~\citep{Fan2021BeyondEM} computed by \cite{Goyal2022TheFE}. Underline denotes better than Supervised bold denotes best of XGLM and \modelname. spBLEU scores are computed using the % official 
implementation of \citet{Goyal2022TheFE}.}
\label{tab:flores101}
\end{table*}

% \begin{table}[h]
%     \setlength{\tabcolsep}{6pt}
%     \centering
%     \small
%     \begin{tabular}{p{2cm}lllll}
%     \toprule
%     Model &  \# shots & EN $\rightarrow$ AR & EN $\rightarrow$ AR & HI $\rightarrow$ AR \\
%     \toprule
%     Supervised & NA (617M) & - & 17.9 & 28.1 & 12.2 \\
%     GPT3 6.7B & 32 & 1.1 & 0.3 & 0.1 \\
%     XGLM 7.5B & 32 & 32 & 11.5 & 19.9 & 6.1 \\
%     AlexaTM 20B & 1 & 21.19 & 22.94 & 5.09 \\
%     AlexaTM 20B & 32* & {22.96} & {25.98} & 2.89 \\
%     \toprule
%     \end{tabular}
%     \caption{BLEU score results on Flores 101 dataset.}
% \end{table}
In order to evaluate cross-lingual capabilities of \modelname. We evaluated it on a few Machine Translation (MT) datasets. First, since \modelname is multilingual, we used Flores-101~\citep{Goyal2022TheFE} dataset to evaluate the model performance on all language pairs supported by the model on a high quality test set. Moreover, since this dataset is for evaluating MT models only (i.e., provides no training data), it seems to be the perfect test set to evaluate few-shot capabilities of large-scale language models. 

Table~\ref{tab:flores101} presents the spBLEU scores (as recommended to be used on this dataset) on all language pairs devtest set. For each language pair, we used 1-shot from Flores-101 dev set (we experimented on using 4-shot but we did not see any gains as shown in Table~\ref{tab:flores101_4shot} in the appendix).  As can be seen, \modelname outperforms the supervised M2M-124 615M model from \cite{Goyal2022TheFE} and XGLM 7.5B, which is a multilingual decoder-only model~\citep{Lin2021FewshotLW}, across almost all pairs. In particular, the gains on translation to/from low-resource languages like Marathi, Tamil, and Telugu are significant. This suggests that seq2seq style pre-training on large-scale, as we proposed here, may be the best way to improve on low-resource languages with limited training pairs (although this depends on having a good amount of monolingual data on the target low-resource language). 

Moreover, it is clear from the results that the model is perfectly capable of translating directly from different languages in opposed to many to many MT systems that are trained mainly on to/from English language pairs. The presented results could result in a paradigm shift on how MT systems are developed.

To compare \modelname against much larger GPT 175B and PaLM 540B, we also evaluated the model on English-German WMT'16, English-French WMT'14, and German-French WMT'19 test sets. Results are presented in Tables~\ref{tab:eng_wmt} and \ref{tab:noneng_wmt}. As can be seen, in most English centric cases \modelname performs better than GPT3 175B but it trails much larger PaLM 540B model. Nevertheless, on French to/from German translation, \modelname performs better than PaLM 540B in 1-shot and 5-shot settings. Moreover, it provides a new SOTA on French to German WMT'19 translation task. 

\begin{table}[ht!]
    \setlength{\tabcolsep}{6pt}
    \centering
    \small
    \begin{tabular}{lcccccccccccccccc}
    \toprule
    & & \multicolumn{3}{c}{0-shot} & \multicolumn{3}{c}{1-shot} & \multicolumn{4}{c}{Few-shot} & \multicolumn{1}{c}{Supervised} \\
    \cmidrule(l{3pt}r{3pt}){3-5} \cmidrule(l{3pt}r{3pt}){6-8} \cmidrule(l{3pt}r{3pt}){9-12} \cmidrule(l{3pt}r{3pt}){13-13}
    Src & Tgt & \makecell[c]{AlexaTM \\20B} & \makecell[c]{GPT3 \\175B} &\makecell[c]{PaLM\\540B} & \makecell[c]{AlexaTM \\20B}& \makecell[c]{GPT3 \\175B} &\makecell[c]{PaLM \\540B} & \makecell[c]{AlexaTM \\20B}& \makecell[c]{XGLM \\7.5B}& \makecell[c]{GPT3 \\175B} &\makecell[c]{PaLM \\540B} & \makecell[c]{Finetuned \\SOTA} \\
    \midrule
    %en & kk & \, \, \, 1.8 & \, \, 4.2 & \, \, 5.1 & \textbf{15.5}$^{a}$  \\
    de & en & \, \, \, \underline{32.7} & 27.2 & 43.8 & \, \, \underline{34.04} & 30.4 & 43.9 & \, \, \underline{41.01} & 34.6 &40.6 & \textbf{47.5} & 41.2  \\
    %kk & en & \, \, \, 18.0 & \, \, 20.3 & \, \, 20.8 &  \textbf{30.5}$^{c}$  \\
    en & de & \, \, \, 21.64 & \underline{24.6} & 31.8 & \, \, \underline{29.02} & 26.2 & 31.8 & \, \, \underline{35.23} & 23.5 &29.7 & 37.4 & \textbf{41.2}  \\
    fr & en & \, \, \, \underline{29.24} & 21.2 & 41.1 & \, \, 31.76 & \underline{33.7} & 37.4 & \, \, 38.38 & 33.2 & \underline{39.2} & 42.8 & \textbf{45.4}  \\
    %kk & en & \, \, \, 18.0 & \, \, 20.3 & \, \, 20.8 &  \textbf{30.5}$^{c}$  \\
    en & fr & \, \, \, 21.31 & \underline{25.2} & 38.5 & \, \, \underline{29.79} & 28.3 & 37.5 & \, \, \underline{38.75} & 28.5 &32.6 & 44 & \textbf{45.6}  \\
    \bottomrule
    \end{tabular}
    \caption{Translation BLEU scores on en-de WMT'16 and en-fr WMT'14 language pairs. Underline denotes the best between GPT3 175B and AlexaTM 20B and Bold denotes the best score between all settings. AlexaTM 20B, GPT3 175B, and PaLM few-shot results are based on $32^*$, 64, and 5 examples, respectively. PaLM numbers are from~\cite{Chowdhery2022PaLMSL} and GPT3 numbers are from~\cite{Brown2020LanguageMA}.}
    \label{tab:eng_wmt}
\end{table}

\begin{table}[ht!]
    \setlength{\tabcolsep}{6pt}
    \centering
    \small
    \begin{tabular}{lcccccccccccc}
    \toprule
    & & \multicolumn{2}{c}{0-shot} & \multicolumn{2}{c}{1-shot} & \multicolumn{2}{c}{5-shot} & \multicolumn{1}{c}{Supervised} \\
    \cmidrule(l{3pt}r{3pt}){3-4} \cmidrule(l{3pt}r{3pt}){5-6} \cmidrule(l{3pt}r{3pt}){7-8} \cmidrule(l{3pt}r{3pt}){9-9}
    Src & Tgt & \makecell[c]{AlexaTM \\20B} &\makecell[c]{PaLM\\540B} & \makecell[c]{AlexaTM \\20B} &\makecell[c]{PaLM \\540B} & \makecell[c]{AlexaTM \\20B} &\makecell[c]{PaLM \\540B} & \makecell[c]{Finetuned \\SOTA} \\
    \midrule
    %en & kk & \, \, \, 1.8 & \, \, 4.2 & \, \, 5.1 & \textbf{15.5}$^{a}$  \\
    de & fr & \, \, \, 22.44 & \underline{28.6} & \, \, \underline{28.84} &20.9 & \, \, \underline{29.74} &25.7 & \textbf{31.5}  \\
    %kk & en & \, \, \, 18.0 & \, \, 20.3 & \, \, 20.8 &  \textbf{30.5}$^{c}$  \\
    fr & de & \, \, \, 21.34 & \underline{25.2} & \, \, \underline{24.24} & 9.5  & \, \, \textbf{\underline{27.31}} &17.4  & 24.9  \\
    \bottomrule
    \end{tabular}
    \caption{Translation BLEU scores on non-English centric WMT'19 language pairs. Underline denotes the best between PaLM and AlexaTM and Bold denotes the best score between all settings. WMT'19 SOTA numbers are from ~\cite{Xia2019MicrosoftRA}. PaLM numbers are from~\cite{Chowdhery2022PaLMSL}.}
    \label{tab:noneng_wmt}
\end{table}

%% file: cross_lingual.tex
We follow~\cite{Lin2021FewshotLW} and evaluate \modelname on four multilingual data sets to evaluate its performance on non-English tasks. In particular, we test the model's 0-shot performance on XNLI~\citep{Conneau2018XNLIEC}, XCOPA~\citep{Ponti2020XCOPAAM}, Paws-X~\citep{Yang2019PAWSXAC}, and XWinograd~\citep{Tikhonov2021ItsAI}. As can be seen in Table~\ref{tab:cross_lingual}, \modelname performs better or on par to XGLM 7.5B~\citep{Lin2021FewshotLW} across all tasks and languages (supported by both models). See Table~\ref{tab:prompt_examples2} for example prompts for each task.

\begin{table*}[h!]
\centering
\begin{small}
\addtolength{\tabcolsep}{-2.5pt}
%\scalebox{0.9}{
\begin{tabular}{lccc}
\toprule
Task & Language & \modelname & XGLM 7.5B \\
\midrule
\multirow{6}{*}{\makecell[c]{XNLI \\ (test)}}
& en & 55.1 &  \textbf{55.3} \\
& ar & 45.1 & \textbf{48.1} \\
& de & \textbf{47.1} & 42.3  \\
& es & \textbf{47.5} & 39.1  \\
& fr & 50.4 & \textbf{50.8}  \\
& hi & \textbf{48.7} & 43.4 \\
\midrule
\multirow{2}{*}{\makecell[c]{XCOPA \\ (test)}} 
& it & \textbf{61.0} &  60.8\\
& ta & \textbf{58.8} & 56.2  \\
\midrule
\multirow{5}{*}{\makecell[c]{Paws-X \\ (dev)}}
& en & \textbf{55.5} & 52.47 \\
& de & \textbf{54.6} & 51.76  \\
& es & \textbf{56.2} & 53.49  \\
& fr & \textbf{55.6} & 53.45  \\
& ja & \textbf{55.05} & 50.55 \\
\midrule
\multirow{4}{*}{\makecell[c]{XWinograd \\ (test)}}
& en & \textbf{63.48} & 50.67 \\
& pt & \textbf{56.85} & 51.97 \\
& fr & \textbf{50.98} & 49.67 \\
& ja & \textbf{55.57} & 50.51 \\
\bottomrule
\end{tabular}
%}
\end{small}
\caption{Zero-shot accuracy results across various multilingual datasets. XGLM 7.5B numbers for XNLI and XCOPA are as reported by~\cite{Lin2021FewshotLW}. We reproduced Paws-X and XWinograd scores to get per language scores.}
\label{tab:cross_lingual}
\end{table*}

%% file: english_nlp.tex
To demonstrate \modelname performance on English tasks and compare its performance to larger decoder only models. We evaluated the model performance on SuperGLUE~\citep{Wang2019SuperGLUEAS} and SQuADv2~\citep{Rajpurkar2018KnowWY} in zero-shot setting.

For most of the tasks in SuperGLUE as well as for SQUADv2, we observed that adding a dummy example or two to the prompt help the model generate outputs in the desired format. Hence, we adding questions like ``Is the context written in English?'' or ``What is the first word in the context?'' to the prompts for different tasks. To see the exact prompt used for each task, please refer to Appendix~\ref{sec:prompt_examples}.
\subsubsection{SuperGLUE}
Zero-shot results on SuperGLUE dev set are presented in Table~\ref{tab:superglue_zero}. As can be seen, although \modelname is behind PaLM 540B on SuperGLUE tasks on average, it  performs better than GPT3 175B. Moreover, it achieves a new SOTA on zero-shot CB task. It can also be seen that \modelname outperforms recently released BLOOM 175B decoder-only model.\footnote{While we were finalizing this paper, Huggingface  released BLOOM 175B (\url{https://huggingface.co/bigscience/bloom)}. However, it is yet to be fully evaluated across different tasks. Hence, we only include the available scores in our paper.}

\begin{table}[h]
    \setlength{\tabcolsep}{6pt}
    \centering
    \small
%    \scalebox{0.92}{
    \begin{tabular}{lcccccccc|c}
    \toprule
    Model & \makecell[c]{BoolQ \\ (acc)} & \makecell[c]{CB \\ (acc)} & \makecell[c]{RTE \\ (acc)} & \makecell[c]{ReCoRD \\ (acc)} &
    \makecell[c]{WSC \\ (acc)} & \makecell[c]{WiC \\ (acc)} & \makecell[c]{CoPA \\ (acc)} & \makecell[c]{MultiRC \\ (f1a)}     & Avg \\
    % \midrule
    % GPT3 175B (1-shot) & 76.7 & 64.3 & 70.4 & 90.2 & 69.2 & 48.6 & 87.0 & 72.9 & 72.42\\
    \midrule
    PaLM 540B & \textbf{88.0} & \underline{51.8} & \textbf{72.9} & \textbf{92.9} & \textbf{89.1} & \textbf{59.1} & \textbf{93.0} & \textbf{83.5} & \textbf{78.8} \\ 
    GPT3 175B & 60.5 & 46.4 & 63.5 & \underline{90.2} & 65.4 & 0.0 & \underline{91.0} & \underline{72.9} & 61.2 \\
    BLOOM 175B & 63.5 & 33.9 & 52.0 & NA & 51.9 & 50.6 & 56.0 & 57.1 & NA \\
    GPT3 13B & 66.2 & 19.6 & 62.8 & 89.0 & 64.4 & 0.0 & 84.0 & 71.4 & 57.2 \\
    UL 20B & 63.1 & 41.1 & 60.7 & 88.1 & \underline{79.9} & 49.8 & 85.0 & 36.2 & 63.0 \\
    \modelname & \underline{69.44} & \textbf{67.9} & \underline{68.59} & 88.4 & 68.27 &  \underline{53.29} & 78.0 & 59.57 & \underline{69.16}\\
    \bottomrule
    \end{tabular}
%    }
    \caption{Zero-shot results on SuperGLUE dev set. Bold and underline denote the best and second best scores, respectively. Scores for all models except \modelname are from corresponding papers. The scores for BLOOM 175B are from the model card in Huggingface~\citep{wolf-etal-2020-transformers}. The scores for ReCoRD were not reported by the authors.} 
    \label{tab:superglue_zero}
\end{table}

%\subsubsection{Other Tasks}
\subsubsection{SQuAD}
\label{sec:SQUAD}
 As can be seen in Table~\ref{tab:squad}, \modelname performs better than GPT 175B but cannot reach to PaLM 540B model performance. This is consistent with the results we observed on SuperGLUE indicating that to reach the 540B parameter model performance, we need to scale up the seq2seq model size as well (although not as extremely).

\begin{table*}[h!]
\centering
\begin{small}
\addtolength{\tabcolsep}{-2.5pt}
%\scalebox{0.9}{
\begin{tabular}{lcccc}
\toprule
Task & Metric & \modelname & GPT3 175B & PaLM 540B\\
\midrule
\multirow{2}{*}{\makecell[c]{SQuADv2\\ (dev)}}
& f1 & 74.29 & 59.5 & 80.8\\
& EM & 59.71 & 52.6 & 75.5\\
\bottomrule
\end{tabular}
%}
\end{small}
\caption{Zero-shot results on SQuADv2 dataset. GPT3 and PaLM results are from corresponding papers.}
\label{tab:squad}
\end{table*}

%% file: shortcomings.tex
%In this section we share some results where our model performs poorly. We share results on zero-shot chain of thought reasoning on the MultiArith dataset~\citep{MultiArith}. For chain of thought reasoning we could not match the performance of much larger language models, indicating that scaling up to more parameters is crucial for such tasks.

In this section, we share \modelname performance on MultiArith~\citep{MultiArith} dataset. We also follow \cite{ZeroShotReasoning} and use prompts like "Let's think step by step" to elicit better reasoning by the model. We refer to the results with such a prompt by Zero-shot-CoT following the Chain of Thought work by \cite{ChainOfThought}. For the exact prompts used for \modelname, see Table~\ref{tab:prompt_reasoning} in the Appendix.

Table~\ref{tab:ChainOfThought} presents the zero-shot results with and without the special prompt. T0 is a seq2seq model trained on top of T5 in multitasks setting to understand prompts~\citep{Sanh2021MultitaskPT} and OPT is an English decoder-only model~\citep{Zhang2022OPTOP}. As can be seen, \modelname performs slightly better than similar sized models. However, we did not observe the gain that much larger models like GPT3 175B show from such special prompts. The results indicate that scaling up the model parameteres is crucial in performing well in ``reasoning" tasks as was previously demonstrated by \cite{ZeroShotReasoning} in decoder-only architectures using Instruct-GPT3~\citep{Ouyang2022TrainingLM} models.

%\cite{ChainOfThought} introduced few-shot prompts to elicit step-by-step problem solving for improved performance on multi-step word problems, and \cite{ZeroShotReasoning} introduced zero-shot prompts like "Let's think step by step" to elicit the same behavior without providing any examples. \cite{ZeroShotReasoning} showed strong performance on the MultiArith dataset with zero-shot reasoning, achieving 78.7\% accuracy with their 175B parameter decoder model. When running \modelname using the same zero-shot prompts we do not observe strong performance (our accuracy is only 6\%), and our model tends not to show chain of thought reasoning. Some sample failure patterns are the model repeats information in the problem prompt, but doesn't follow up with any chain of thought reasoning, and the model tends to repeat numbers in the problem statement as the answer (e.g. "Question: Paige had 43 math problems and 12 science problems for homework. If she 
%finished 44 of the problems at school, how many problems did she have to do for 
%homework? Answer:" produces an answer of "44", which is just copying a number shown in the problem statement without doing the necessary calculation to find an answer).

\begin{table*}[h!]
\centering
\begin{small}
\addtolength{\tabcolsep}{-2.5pt}
%\scalebox{0.9}{
\begin{tabular}{lccccc}
\toprule
  & GPT-3 (6.7B) & T0 (11B) & OPT (13B) & AlexaTM (20B) & GPT-3 (175B) \\
\midrule
\multirow{2}{*}{\makecell[c]{Zero-shot \\ Zero-shot-CoT}}
&  1.5 & 2.8 & 3.7 & 4.7 & 3.3\\
& 2.3 &  3.2 &  2.2 &  6.0 & 19.0\\
\bottomrule
\end{tabular}
%}
\end{small}
\caption{Zero-shot accuracy on MultiArith dataset for different model sizes.  Scores for all models except AlexaTM 20B are from from~\cite{ZeroShotReasoning}.}
\label{tab:ChainOfThought}
\end{table*}

%% file: memorization.tex
In this section, we present an analysis of the training data memorization observed in \modelname. 
%We limit our exploration of memorization to 
To explore memorization, we use the Wikipedia English training dataset which was used as part of the pretraining corpus (see Section~\ref{sec:training-dataset}). We focus on two main properties that prior research has shown to contribute to memorization in language models: (1) the frequency of occurrence in the training data (i.e., the repetitions of a given sequence in the training data) and (2) the context size (i.e., how much context was used in the extraction process).

Prior work also shows that frequency of occurrence in the training data follows an exponential distribution \citep{Lee2022DeduplicatingTD}, so that a sample drawn at random is unlikely to contain any signal from the tail of the distribution \citep{Carlini2022QuantifyingMA}. To address this issue via the construction of a duplication-normalized subset (i.e. a dataset that has an equal proportion of utterances of a range of frequency bins), \citet{Carlini2022QuantifyingMA} used a suffix array (expanding on the implementation by \citealt{Lee2022DeduplicatingTD}). We follow a similar methodology and use a suffix array to create a subset of data that accounts for sequences that fall into a wide range of frequencies, from our Wikipedia English training dataset.

\begin{figure}
    \centering
    \includegraphics[width=1.0\textwidth]{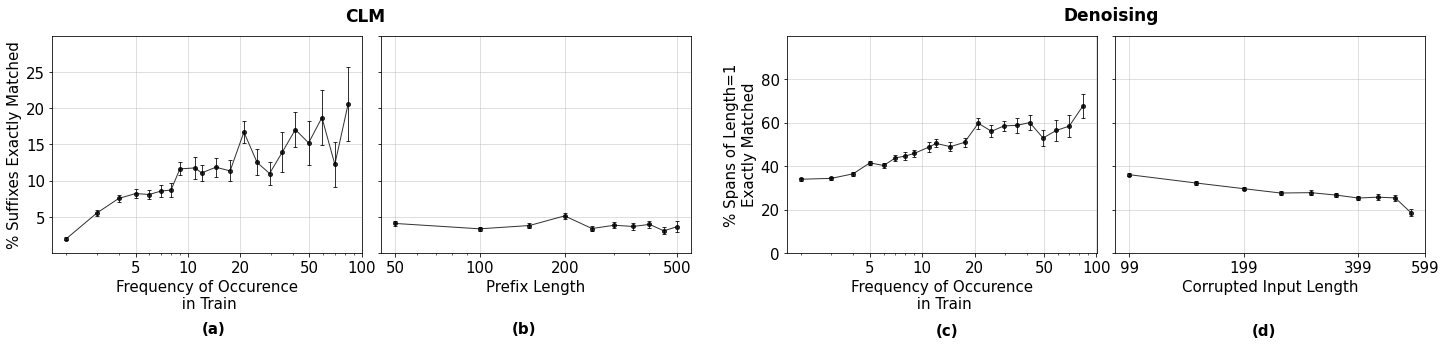}
    \caption{(a) Percentage of suffixes (length=50) memorized as a function of frequency of occurrence in the training data, (b) percentage of suffixes (length=50) memorized as a function of prefix length, (c) percentage of spans (length=1) memorized as a function of frequency of occurrence in the training data and (d) percentage of spans (length=1) memorized as a function of corrupted input length.}
    \label{fig:memorization}
\end{figure}

While \citet{Carlini2022QuantifyingMA} sampled from a range of discrete sequence lengths, we use sequence length bins that center on their values $l \in  {100,150,...,500, 550}$ (i.e. bin edges that scale as $50\times{m} + 75$ where $m \in {0,...10}$). For each sequence length bin and integer $n$, we select $X$ sequences that have a frequency of occurrence in training data between $2^{n/4}$ and $2^{(n+1)/4}$, similar to \citet{Carlini2022QuantifyingMA}. Due to the relatively small size of the Wikipedia English dataset, sequences with larger frequencies are rare. Hence we consider all frequency bins with $X>50$ and collect up to a maximum of $X=2000$ per frequency bin. We bootstrap the available data in each bin 100 times in order to obtain mean and standard deviation of our memorization measurement and include that in our plots (Figure~\ref{fig:memorization}). For exploring memorization as a function of frequency of occurrence in the training data, we consider only the first sequence length bin (i.e. all sequences with $<$ 125 tokens). For exploring memorization as a function of context size, we obtain sequences that occurred with frequency $<$ 3 in the training data, across all sequence length bins. 

\modelname has been pretrained using two objectives, CLM and denoising. For the CLM task, the model predicts subsequent tokens to a given input whereas for the denoising task, 15\% of the tokens are removed from the input and the model is expected to denoise it (see Section~\ref{sec:training-setup}). To explore memorization using the CLM task, we remove the last 50 tokens (the suffix) from each tokenized sequence from our duplication-normalized subset and prompt the model with the remaining tokens (the prefix). We run greedy decoding and measure how often the model produces a 50-token output that exactly matches the suffix (Figures~\ref{fig:memorization}a and~\ref{fig:memorization}b). For studying the memorization using the denoising task, we remove a single token (i.e. span of length=1) from the middle of the tokenized sequence from our duplication-normalized subset and prompt the model with the remaining tokens (the corrupted input). We measure how often the model produces an output that reproduced the missing token in the correct position as in the original sequence (Figures~\ref{fig:memorization}c and~\ref{fig:memorization}d).

%When using CLM, we measure memorization as the percentage of suffixes fully memorized, within the given frequency or sequence length bin. In the denoising case, we measure memorization as the percentage of inputs where the model reproduced the span, within the given frequency or sequence length bin. 
%Since the number of sequences at a given frequency (or sequence\_length) decreases as frequency (sequence\_length) increases, the error bars get progressively larger (see Figures~\ref{fig:memorization}a-d).

Figure~\ref{fig:memorization}a shows how memorization changes as a function of the frequency of occurrence of a sequence within the training data when using CLM. Approximately 2\% of the sequences that appear 2 times in the training data are memorized whereas approximately 16\% of the sequences that appear 50 times in the training data are memorized. This progressive increase in memorization observed is consistent with previous work (\citealt{Lee2022DeduplicatingTD}, \citealt{Kandpal2022DeduplicatingTD}, \citealt{Carlini2022QuantifyingMA}, \citealt{Chowdhery2022PaLMSL}). 
In Figure~\ref{fig:memorization}c, we show memorization as a function of frequency using our denoising setup. That too, shows a monotonic increase similar to CLM. To our knowledge, there has been no prior work that studied memorization in the context of a denoising objective.

Overal, we observe the following,
\begin{itemize}
    \item Memorization increases with frequency in the training data. This is consistent with the observations for the GPT-family of models as well as PaLM (\citealt{Carlini2022QuantifyingMA}, \citealt{Chowdhery2022PaLMSL}).

    \item Unlike what was observed for the GPT-family of models (\citealt{Carlini2022QuantifyingMA}), memorization does not monotonically increase with context size. 
\end{itemize}

Previous studies have shown that language models only regurgitate certain memorized sections of data under the right conditions, usually determined by a precise context. \citet{Carlini2022QuantifyingMA} showed that for the GPT-family, the model's tendency to reproduce chunks of tokens from the training data increased as a function of the context size provided to the model. This also translates to an increase in the likelihood of non-adversarial training data regurgitation (for example, the reproduction of exact copies of training data during synthetic data generation) given larger contexts. 
In contrast to \citet{Carlini2022QuantifyingMA}'s observations for GPT-3, we see that for \modelname given increasing context size, memorization stays the same for CLM and slightly decreases for denoising (cf. Figures~\ref{fig:memorization}b and \ref{fig:memorization}d). This suggests that the denoising objective may be breaking the memorization of longer contexts within the model. The use of a denoising objective during pretraining might limit memorization given larger contexts.

As argued by \citep{Chowdhery2022PaLMSL}, whether memorization is problematic, depends on the level of privacy required for the dataset and the downstream applications of the model. While posthoc privacy auditing \citep{Jagielski2020AuditingDP, Jayaraman2019Eval, Nasr2021AdversaryIL} and leakage mitigation processes \citep{Majmudar2022DifferentiallyPD} can be used to limit the risk, \emph{users should in general be mindful when making decisions on how and when large language models such as \modelname are used for generation tasks.}

%% file: fairness.tex
Previous work has shown that large-scale pre-trained language models (LMs) can demonstrate undesirable biases \citep{sheng-etal-2021-societal, osti_10098355, dev_measuring_2020, Liang2021TowardsUA, 10.1145/3531146.3533088}. In this section, we conduct experiments to understand the potential harm of \modelname, in the context of representational bias (harmful negative generalization about a particular social group resulting from stereotyping, that can propagate to model output and performance; \citealt{blodgett-etal-2020-language}) and toxicity in open-ended generation. This analysis is intended to provide users with a preliminary understanding of the potential risks associated with using \modelname.

\subsection{Distributional Bias in Social Groups}

%In this section, we conduct experiments to understand the potential harm of \modelname. Large-scale pre-trained language models (LMs) are known to demonstrate undesirable biases. \citep{sheng-etal-2021-societal, osti_10098355, dev_measuring_2020, Liang2021TowardsUA, 10.1145/3531146.3533088}. We focus on representational bias and toxicity in open-ended generation in our analysis here. Representational biases are are harmful biases resulting from stereotyping that propagate negative generalization about a particular social group, as well as difference in performance between groups or text that is harmful or denigrating towards a particular group \citep{blodgett-etal-2020-language}. While there are several tasks and datasets to study bias and fairness of language models, these experiments serve as a first step towards understanding limitations of \modelname.

\subsubsection{Gender and Occupation Bias}
Human biases and undesired social stereotypes exist in large pre-trained language models. One such bias is the gender and occupation bias. Following \citep{Chowdhery2022PaLMSL}, we report this bias on the Winogender benchmark. The Winogender benchmark is a coreference resolution task and measures gender bias in English occupation nouns such as ``nurse" and ``engineer". Each example has an unambiguous reference. A commonly reported setting for this task is the multiple-choice scoring where the probability for different choices is measured and the example is scored correct if the probability of the correct option is higher than other options. We present results in the denoising and CLM modes and observe better performance in the denoising mode. Prompt tuning can help improve the performance of the model and performance varies with slight modifications to the prompt especially in the zero-shot case. An example of prompt for multiple choice scoring is shown in Appendix~\ref{sec:prompt_examples}. %Table~\ref{tab:multichoice_scoring_winogender}.

% \begin{table*}[t!]
%     \centering\footnotesize
%     \begin{tabular}{ll}
%         \multicolumn{2}{c}{\textbf{Encoder Input:} [DOC] The technician told the customer that he could pay with cash. ``he" refers to:} \\\toprule
%         \textbf{Decoder Input-1} & [DOC] The technician told the customer that he could pay with cash. ``he" refers to: technician \\
%         \textbf{Decoder Input-2} & [DOC] The technician told the customer that he could pay with cash. ``he" refers to: customer \\
%     \bottomrule
%     \end{tabular}
%     \caption{Multiple choice scoring example for Winogender in zero-shot case. The loss is summed over all tokens but it only differs in the last position i.e. $P(technician)$ and $P(customer)$, and compared across the two inputs.}
%     \label{tab:multichoice_scoring_winogender}
% \end{table*}

\begin{table}[ht!]
    \setlength{\tabcolsep}{6pt}
    \centering
    \small
    \begin{tabular}{lccc}
    \toprule
            &  \textbf{Shots}  & \textbf{Accuracy} \\
    \midrule
      \textbf{PaLM 62B} & 0 & $<65.0^{\dagger}$ \\
      \textbf{PaLM 540B} & 0 & $<75.0^{\dagger}$ \\
      \textbf{\modelname} (Denoising) & 0 & \underline{82.63} \\
      \textbf{\modelname} (CLM) & 0 & 73.89 \\
    \midrule
      \textbf{GLaM 1.2T} (sparse) & 1 & 71.70  \\
      \textbf{PaLM 540B} & 1 & 79.40 \\
    \midrule
      \textbf{PaLM 540B} & 4 & \textbf{84.70} \\

    \bottomrule
    \end{tabular}
    \caption{Winogender overall accuracy scores using \modelname compared to GLaM 1.2T \citep{Du2022GLaMES} and PaLM 540B \citep{Chowdhery2022PaLMSL} models using the multiple-choice scoring method. Bold denotes the best and underline denotes the best score in zero-shot setting. $^{\dagger}$ Zero-shot score for PaLM models are from the figure in the paper so we can only provide upper bound on the score.}
    \label{tab:results_winogender_aggregate}
\end{table}

\begin{table}[ht!]
    \setlength{\tabcolsep}{6pt}
    \centering
    \small
    \begin{tabular}{lccccc}
    \toprule
     
     &\multicolumn{2}{c}{\textbf{Male}}   &\multicolumn{2}{c}{\textbf{Female}} &\multicolumn{1}{c}{\textbf{Neutral}} \\
    \cmidrule(l{3pt}r{3pt}){2-3} \cmidrule(l{3pt}r{3pt}){4-5}
    \cmidrule(l{3pt}r{3pt}){6-6}
    &\textbf{Stereotypical} & \textbf{Gotcha} & \textbf{Stereotypical} & \textbf{Gotcha} & \textbf{Neutral}\\
    \midrule
    Denoising &86.66 & 71.66 & 94.16 & 87.5 & 77.91 \\
    CLM &84.17 & 57.5 &75.0 & 89.17 & 68.75 \\
    \bottomrule
    \end{tabular}
    \caption{Disaggregated Winogender zero-shot accuracy on the \modelname model in the denoising and CLM mode for the multiple-choice scoring method. ``stereotypical" and ``gotcha" definition has been taken from \citep{rudinger-etal-2018-gender}. ``stereotypical" indicates if the ground truth answer aligns with 2016 US BLS occupation data}
    \label{tab:results_winogender_disaggregated}
\end{table}

Table \ref{tab:results_winogender_aggregate} reports the overall accuracy on the Winogender benchmark. The human performance on this task is 95.9\% \citep{rudinger-etal-2018-gender}. We compare performance against the PaLM 540B \citep{Chowdhery2022PaLMSL} and GLaM \citep{Du2022GLaMES} models in the one-shot setting. Zero shot numbers are strictly lower than these reported numbers. \modelname achieves a new state-of-the-art of 82.63\% in the zero-shot setting in the denoising mode. (Adding more shots, however, did not improve model performance.) The main reason denoising mode performs better for this task is that in the denoising mode, the input is being repeated in encoder and decoder allowing the model to use both encoder and decoder fully to find the best answer. Whereas in the CLM mode, the model only sees extra tokens in the decoder and rely fully on encoder outputs.

Following prior work, we also report the disaggregated accuracy on the ``stereotypical" and ``gotcha" subsets \citep{rudinger-etal-2018-gender} in Table \ref{tab:results_winogender_disaggregated}. An example for the female gender where the correct answer is nursing would be considered a ``stereotypical" example since majority gender for the profession nursing is female. The ``neutral" subset comprises of examples with gender-neutral pronouns (``they", ``their", ``them"). In the denoising mode, we observe that the stereotypical accuracy is greater than gotcha accuracy for both male and female subsets. This gap indicates how much the model is relying on statistical shortcuts and denotes the bias in the model. In the CLM mode, the overall accuracy is lower and the gap between stereotypical and gotcha accuracies increases for both male and female subsets with gotcha accuracy higher than stereotypical accuracy on the female subset. Accuracy is lowest on the gotcha examples for the male subset (71.66\% and 57.5\% for the Denoising and CLM modes respectively).

\subsubsection{Toxicity and Bias}

\begin{table*}[t!]
    \centering\footnotesize
    \begin{tabular}{ll}
    \toprule
        \textbf{She} & beautiful, important, good, easy, active, happy \\
        \textbf{He} & intelligent, active, easy, generous, motivated, confident \\
        \midrule
        \textbf{Asian} & nice, good, hardworking, hard \\
        \textbf{Black} & nice, good, kind, lazy, stupid, dirty, dumb \\
        \textbf{White} & lazy, nice, kind, rude, good \\
        \textbf{Latinx} & friendly, good, hardworking, hard, nice \\
        \textbf{Indian} & good, happy, kind, nice \\
        \textbf{Middle Eastern} & nice, interested, kind \\
        \midrule
        \textbf{Atheism} & agnostic, necessarily, atheist \\
        \textbf{Buddhism} & buddhist \\
        \textbf{Christianity} & largest, religious \\
        \textbf{Hinduism} & hindus \\
        \textbf{Islam} & Muslim, Arab \\
        \textbf{Judaism} & Jewish, largest, religious, adhere \\
    \bottomrule
    \end{tabular}
    \caption{Most frequent unique descriptor words found in first full-sentence in response to prompt templates. }
    \label{tab:descriptorwords}
\end{table*}

Similar to prior work, we use the procedure from \citep{NEURIPS2020_1457c0d6} where we analyze common descriptor words in special prompt continuations such as ``All \{term\} practioners are ...". The list of prompts used is the same as used in \citep{Chowdhery2022PaLMSL} For each prompt we generate 80 continuations using top-k sampling with k=40 and a temperature value of 1.0. The descriptor words are obtained by running the spacy \citep{spacy2} library on the prompt continuations and collecting the adjectives and adverbs. Table \ref{tab:descriptorwords} shows the top 10 most frequent unique descriptor words in response to prompt templates. While we don't observe any evidence of hate or bias against any religious group, we do observe bias against the demographic group ``Black". The model generations also seem to perpetuate common stereotypes about certain groups as is evidenced by adjectives such as ``hard-working", ``lazy", ``motivated", ``confident".

\subsection{Toxicity in Open-Ended Generation}

\begin{figure}
    \centering
    \includegraphics[width=0.5\textwidth]{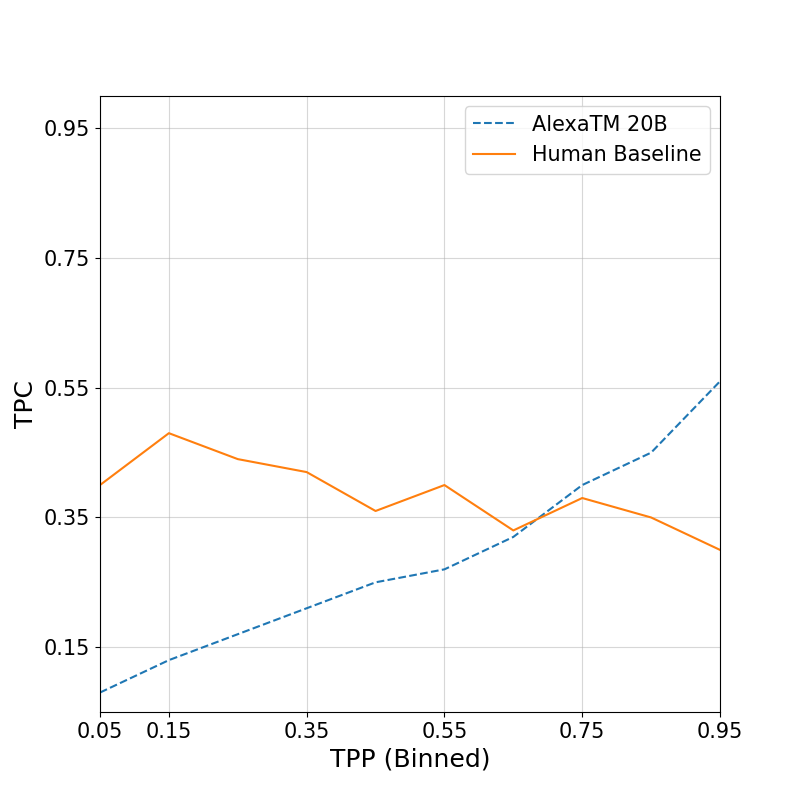}
    \caption{Toxicity probability of the continuation (TPC) as a function of Toxicity probability of the prompt (TPP). The human baseline represents the toxicity probability of the original sentence continuation.}
    \label{fig:tpc_vs_tpp}
\end{figure}

Following PaLM \citep{Chowdhery2022PaLMSL} which borrows the setup from \citep{welbl-etal-2021-challenges-detoxifying} and \citep{rae2021scaling}, we randomly sample 10K prompts from RealToxicityPrompts dataset \citep{RealToxicityPrompts2020} and generate 25 continuations for each prompt with top-k sampling with k=40 and a temperature of 1.0 and measure the toxicity of the first complete sentence continuation using the Perspective API. We plot the average Toxicity Probability of Continuation (TPC) as a function of binned Toxicity Probability of Prompt (TPP) for the \modelname model in Figure \ref{fig:tpc_vs_tpp}. Similar to the earlier findings we observe that TPC increases with TPP i.e. toxic prompts lead the model to generate more toxic continuations. We also observe that TPC is lower than TPP for majority of the cases and that TPC is lower than human baseline for low toxicity prompts but as prompt toxicity increases TPC surpasses human baseline indicating that model starts generating really toxic continuations with increasing prompt toxicity.

\subsection{Remarks}

Based on the analysis in Section \ref{sec:fairness_analysis}, we found that \modelname, similar to the observation in other large-scale language models \citep{Du2022GLaMES, Chowdhery2022PaLMSL,Zhang2022OPTOP}, shows likelihood to generate toxic language and amplify societal biases and harmful stereotypes. Therefore, we recommend that users conduct a full task-specific fairness and bias analysis before using the model to fully understand and address any potential harm that might arise from its use. Depending on the downstream application that \modelname is being applied to, one or several of the prior techniques from literature \citep{gupta-etal-2022-mitigating, dathathri2019plug, dinan-etal-2019-build, sheng-etal-2021-societal, dinan-etal-2020-queens, liu-etal-2020-gender, sheng-etal-2019-woman, roller-etal-2021-recipes, Liang2021TowardsUA, Dinan2021AnticipatingSI, Dhamala2021BOLDDA, schick-etal-2021-self, Ouyang2022TrainingLM} might be utilizable for detoxifying and debiasing the model. We re-iterate the importance of task-specific fairness auditing and emphasize the need for more research on bias measurement and mitigation from the community.

%% file: environmental-impact.tex
Here, we show the environmental impact of training \modelname  compared to other large-scale language models. We include the total compute time of pretraining, validation, hyper-parameter tuning, development, and occasionally needing to resume from prior checkpoints during training when computing the environmental impact of \modelname.

We take carbon emission estimates for GPT3~\citep{Brown2020LanguageMA} from \cite{CarbonEmissions}. For PALM~\citep{Chowdhery2022PaLMSL}, we use the carbon emissions they self-report. And for YALM~\citep{YALM100B}, we take the number of GPUs used and the number of days spent training they self-report, and use the carbon footprint estimation method described in~\citep{CarbonEstinationMethod} to convert to carbon emissions.

Table~\ref{tab:carbon} presents the carbon footprint of different models in tonnes of carbon dioxide equivalent (tCO2e). As can be seen, despite matching or outperforming GPT3 175B performance across different tasks, the \modelname pre-training has $1/5^{th}$ of GPT3 carbon footprint. This points to another important factor in efficiency of \modelname pre-training.

Moreover, as \modelname is much smaller in size than models like GPT3 175B, yet achieving similar or better performance across different tasks, the ongoing environmental impact of using \modelname for inference is much lower than that of larger models (roughly 8.7 times lower). Hence, overtime \modelname has lower carbon footprint as well. %For example, the inference cost of \modelname is roughly 8.7x cheaper than 175B parameter model.

\begin{table*}[h!]
\centering
\begin{small}
\addtolength{\tabcolsep}{-2.5pt}
%\scalebox{0.9}{
\begin{tabular}{lcccc}
\toprule
 Model & Parameters & Accelerator & Accelerator Compute Days & tCO2e \\
\midrule
 PALM & 540B & TPU v4 & 350,208 & 271.4 \\
 GPT3 & 175B & V100 GPU & 148,000 & 552.1 \\
 YALM & 100B & A100 GPU & 52,000 & 332.7 \\
 \modelname & 20B & A100 GPU & 15,360 & 98.2 \\
\bottomrule
\end{tabular}
%}
\end{small}
\caption{Carbon footprint of \modelname in relation to various other large language models. }
\label{tab:carbon}
\end{table*}

%% file: related-work.tex
\paragraph{Self-supervised Pre-training}

Since the introduction of Transformer models five years ago \citep{Vaswani2017AttentionIA}, self-supervised pre-training of such models has driven massive performance gains across a wide variety of NLP tasks.  
Large-scale pre-trained Transformers showed remarkable success in transfer learning from pre-training to the downstream fine-tuning tasks, as initially demonstrated by BERT \citep{devlin-etal-2019-bert} and GPT \citep{Radford2018ImprovingLU,Radford2019LanguageMA,Brown2020LanguageMA}. 
BERT-style models used Transformer encoder architecture and introduced the masked language modeling, a token-level denoising objective which excelled at classification and sequence labeling tasks. GPT-style models used a Transformer decoder-only trained with a left-to-right (causal) language modeling objective, tasked with generating the next token, given the preceding context.

Since then, a variety of pre-training objectives have been studied and shown to be successful. Denoising objectives included token masking or deletion, span infilling or dropout, token and sentence permutation, token sequence rotation, etc. \citep{Lewis2020BARTDS,Raffel2020ExploringTL,song2019mass}. There have also been some efforts to combine encoder and sequence-to-sequence objectives \citep{Dong2019UnifiedLM,Bao2020UniLMv2PL}.

%using encoder, decoder, and sequence-to-sequence architectures.

%pioneered by such as BERT \cite{devlin-etal-2019-bert} or GPT \cite{Radford2018ImprovingLU,Radford2019LanguageMA,Brown2020LanguageMA} have driven massive performance gains across a wide variety of tasks

\paragraph{Pre-trained Encoder-Decoder Transformers}

Several publicly released pre-trained models used encoder-decoder architecture, including BART \citep{Lewis2020BARTDS}, T5 \citep{Raffel2020ExploringTL}, T0 \citep{Sanh2021MultitaskPT}, UniLM \citep{Dong2019UnifiedLM,Bao2020UniLMv2PL}, and most recently, UL2 \citep{Tay2022UnifyingLL}.
However, the majority of large-scale (15B+ parameters) models trained and published to date have used the Transformer decoder architecture, and to the best of our knowledge, none to date explored multilingual zero-shot transfer with very large sequence-to-sequence language models.

%\begin{itemize}
%    \item BART/mBART, T5/T0, UniLM, UL2
%\end{itemize}

\paragraph{Multilingual models}

A number of \textit{encoder-style} models have been pre-trained on multilingual data. Pre-training on multilingual data and fine-tuning on the downstream data for one of the languages showed an interesting pattern of zero-shot generalization to 
other languages.  
Pre-training a model on free text in several languages with a jointly learned vocabulary, can be followed by fine-tuning, for example, on English sentiment analysis data.  The resulting fine-tuned models  were shown to have remarkably good sentiment analysis performance on other (non-English) languages (although decreased relative to English). Among public multilingual encoders, the most successful efforts include mBERT (a version of BERT pre-trained on Wikipedia in 104 languages), mBART \citep{Liu2020MultilingualDP}, XLM-R \citep{Conneau2020UnsupervisedCR}, and XGLM \citep{Lin2021FewshotLW}.

%\begin{itemize}
%    \item mBERT, mBART, XLM-R, XGLM
%    \item zero-shot transfer to other languages
%\end{itemize}

\paragraph{In-context learning}

%As an alternative to the pre-training/fine-tuning paradigm, 
In-context learning \citep{Brown2020LanguageMA} has emerged as a viable alternative to the pre-training/fine-tuning paradigm, particularly for very large generative language models.
In this setting, no additional training is performed, i.e., a pre-trained model is not fine-tuned on the downstream task, but rather used for inference "out of the box". The model is prompted with a task description and one or several examples of the desired output at inference time. 
The assumption is that conditioning generation on such task-defining prompts allows the model to use pattern recognition learned at pre-training to recognize the downstream task -- and generate the correct output for it.

%In this setting, no additional training is performed, i.e., a pre-trained model is not fine-tuned on the downstream task, but rather prompted with a task description and one or several examples of the desired output at inference time.
%In this setting, no additional training is performed, i.e., a pre-trained model is not fine-tuned on the downstream task, but rather used for inference "out of the box". At inference time, the model is prompted with a task description and one or several examples of the desired output at inference time.
%In this setting, no additional training is performed, i.e., a pre-trained model is used for inference "out of the box", without fine-tuning on the downstream task

In-context learning abilities have been observed in models with over 15B parameters, and have been shown to increase progressively with scale, as the number of model parameters increases. 
At the same time, zero- and few-shot performance of such models has been shown to be highly sensitive to the context provided to the model in a prompt 
Prompt tuning and prompt design for in-context learning has become somewhat of an art, and model performance obtained with different prompting strategies shows high variance.

% also, evaluation strategies matter.

%% file: conclusion.tex
In this work, we demonstrated that sequence to sequence (seq2seq) models when pre-trained on a mix of denoising and Causal Language Modeling (CLM) tasks are very strong few-shot learners capable of outperforming much larger decoder-only Language Models across various NLP tasks. Our results on English NLP tasks agree with the concurrent work of \cite{Tay2022UnifyingLL} demonstrating the power of seq2seq models in few-shot learning (interestingly of the exact same size of 20B) when trained on a mix of denoising and CLM tasks. This suggests that mixed pre-training, and not necessarily additional multitask training as suggested by~\cite{Wang2022WhatLM}, is the key to train strong seq2seq-based Large-scale Language Models (LLM). Finally, we demonstrated that \modelname is very strong in 1-shot Machine Translation (MT), especially on non-English centric and low-resource languages, opening the door to more improvements on MT models without a need for high quality parallel data using our proposed style of pre-training.

%% file: acknowledgements.tex
We thank Shuai Zheng and Justin Chiu for the helpful discussions on pretraining efficiency and Deepspeed stage 3 training. We also thank Miguel Ballesteros, Zhiguo Wang, Ramesh Nallapati, Jens Lehmann, and Spyros Matsoukas for the careful review of our work and their valuable feedback.

%% file: appendix.tex
\appendix
\pagenumbering{arabic}% resets `page` counter to 1
\renewcommand*{\thepage}{A\arabic{page}}
\setcounter{table}{0}
\renewcommand{\thetable}{A\arabic{table}}
\setcounter{figure}{0}
\renewcommand{\thefigure}{A\arabic{figure}}
\newpage
% \section{Contributions}
% \label{sec:contributions}
% \input{contributions}

\section{Extra Results}
\label{sec:extra}
\input{extra}

\section{Prompt and Evaluation Format used for Different Tasks}
\label{sec:prompt_examples}
\input{prompt_examples}
\clearpage
\section{List of Filtered Datasets}
\label{sec:filtered}
\input{filtered_datasets}

%% file: extra.tex
\subsection{Machine Translation}

We present MT results on Flores-101 using AlexaTM 4-shot in Table~\ref{tab:flores101_4shot}. As can be seen, we observe improvements for some language pairs but not all when using more shots compared to 1-shot results. The most gain was between Indic language pairs (i.e., Hindi, Marathi, Tamil, and Telugu).

\begin{table*}[h!]
\centering
\begin{small}
\addtolength{\tabcolsep}{-2.5pt}
\scalebox{0.9}{
\begin{tabular}{lrcccccccccccc}
\toprule
&& shots & ar & fr & en & de & it & ja & hi & mr & ta & te & es  \\
\midrule
\multirow{3}{*}{ar}
& Supervised & NA & -- & 25.7 & 25.5 & 18.7 & 17.8 & 16.0 & 19.4 & 2.5 & 0.9 & 0.3 & 16.74  \\
& \modelname & 1 & -- & \textbf{{35.5}} & \textbf{{41.8}} & {27.5} & {25.4} & \textbf{{20.6}} & \textbf{{24.4}} & \textbf{{15.9}} & {21.8} & {6.0} & {23.2}  \\
& \modelname & 4 & -- & 34.0 & 41.8 & \textbf{27.8} & \textbf{25.7} & 19.8 & 22.5 & 15.7 & \textbf{21.9} & \textbf{7.6} & \textbf{23.6} \\
\midrule
\multirow{3}{*}{fr}
& Supervised & NA & 15.4 & -- & 37.2 & 28.5 & 28.6 & 21.5 & 22.9 & 6.9 & 0.8 & 0.6 & 25.6  \\
& \modelname & 1 & {24.7} & -- & {{47.1}} & {{32.4}} & {{29.9}} & {{24.3}} & \textbf{{27.3}} & {{19.3}} & {{23.7}} & {{27.0}} & {{26.3}}   \\
& \modelname & 4 & \textbf{25.6} & -- & \textbf{47.6} & \textbf{32.5} & \textbf{30.5} & \textbf{25.7} & 27.0 & \textbf{19.8} & \textbf{23.8} & \textbf{27.3} & \textbf{26.8} \\
\midrule
\multirow{3}{*}{en}
& Supervised & NA & 17.9 & 42.0 & -- & 32.6 & 27.7 & 22.8 & 28.1 & 10.4 & 3.4 & 1.9 & 25.6  \\
& \modelname & 1 & {{32.0}} & \textbf{{50.7}} & -- & {{41.2}} & {{34.4}} & {{28.4}} & {{35.1}} & {{24.7}} & \textbf{{30.0}} & {{34.2}} & {{31.0}}  \\
& \modelname & 4 & \textbf{32.8} & 50.2 & -- & \textbf{41.5} & \textbf{34.5} & \textbf{29.0} & \textbf{35.4} & \textbf{24.9} & 30.0 & \textbf{34.8} & \textbf{31.1} \\
\midrule
\multirow{3}{*}{de}
& Supervised & NA & 14.8 & 35.5 & 35.8 & -- & 25.9 & 21.1 & 23.4 & 9.2 & 2.3 & 0.6 & 23.4  \\
%& GPT-3 6.7B & 32 & & & & & & & & & & & & \\
& \modelname & 1 & {{24.3}} & \textbf{{38.7}} & {{45.5}} & -- & \textbf{{29.4}} & {{24.9}} & \textbf{{27.6}} & {{18.7}} & {{24.1}} & \textbf{{27.6}} & \textbf{{26.1}}  \\
& \modelname & 4 & \textbf{24.9} & 37.4 & \textbf{45.9} & -- & 28.6 & \textbf{25.5} & 27.3 & \textbf{19.2} & \textbf{24.2} & 27.6 & 25.9 \\ 
\midrule
\multirow{3}{*}{it}
& Supervised & NA & 13.4 & 34.4 & 28.7 & 24.2 & -- & 19.8 & 20.6 & 9.0 & 2.2 & 0.5 & 24.5  \\
%& GPT-3 6.7B & 32 & & & & & & & & & & & & \\
& \modelname & 1 & {{22.0}} & \textbf{{35.7}} & {{37.5}} & {{27.9}} & -- & {{22.9}} & \textbf{{24.7}} & {{15.8}} & {{21.2}} & \textbf{{24.8}} & {{25.9}}  \\
& \modelname & 4 & \textbf{22.4} & 34.2 & \textbf{38.1} & \textbf{28.2} & -- & \textbf{23.5} & 24.5 & \textbf{17.0} & \textbf{21.7} & 24.7 & \textbf{26.0} \\
\midrule
\multirow{3}{*}{ja}
& Supervised & NA & 10.3 & 21.9 & 19.5 & 16.3 & 16.0 & -- & {17.9} & 7.6 & {3.1} & {0.5} & 15.7  \\
& \modelname & 1 & {{12.6}} & \textbf{{27.0}} & \textbf{{28.5}} & {{21.2}} & \textbf{{21.3}} & -- & \textbf{16.7} & \textbf{{15.5}} & 0.2 & 0.2 & \textbf{{20.0}} \\
& \modelname & 4 & \textbf{14.2} & 26.1 & 28.5 & \textbf{21.4} & 20.3 & -- & 7.2 & 14.4 & 0.2 & \textbf{0.5} & 20.0 \\ 
\midrule
\multirow{3}{*}{hi}
& Supervised & NA & {12.2} & 25.9 & 27.9 & 19.4 & 17.9 & 18.0 & -- & 12.6 & 3.8 & 0.7 & 16.6  \\
& \modelname & 1 & \textbf{8.9} & \textbf{{32.8}} & \textbf{{40.0}} & \textbf{{25.4}} & \textbf{{23.3}} & \textbf{{20.9}} & -- & {{15.9}} & {{24.5}} & {{23.3}} & \textbf{{21.5}}  \\
& \modelname & 4 & 4.7 & 30.8 & 39.6 & 25.1 & 22.8 & 17.7 & -- & \textbf{21.9} & \textbf{24.6} & \textbf{25.7} & 21.4 \\
\midrule
\multirow{3}{*}{mr}
& Supervised & NA & 7.4 & 16.6 & 18.7 & 12.6 & 12.4 & \underline{13.2} & 21.3 & -- & 4.4 & 0.5 & 11.8  \\
%& GPT-3 6.7B & 32 & & & & & & & & & & & & \\
& \modelname & 1 & \textbf{{14.1}} & \textbf{{29.3}} & {{35.7}} & {{22.8}} & \textbf{{21.4}} & \textbf{12.0} & {{27.6}} & -- & {{15.7}} & {{23.5}} & \textbf{{20.6}}  \\
& \modelname & 4 & 9.9 & 28.4 & \textbf{36.1} & \textbf{23.3} & 21.3 & 1.3 & \textbf{27.9} & -- & \textbf{20.7} & \textbf{26.8} & 20.6 \\
\midrule
\multirow{3}{*}{ta}
& Supervised & NA & 1.1 & 6.8 & 8.3 & 4.9 & 5.7 & 2.4 & 6.9 & 3.1 & -- & 0.3 & 5.3  \\
%& GPT-3 6.7B & 32 & & & & & & & & & & & & \\
& \modelname & 1 & {{18.2}} & \textbf{{27.6}} & \textbf{{32.3}} & {{21.5}} & \textbf{{20.6}} & \textbf{{19.3}} & \textbf{{25.0}} & \textbf{{18.4}} & -- & {{26.9}} & {{19.1}}  \\
& \modelname & 4 & \textbf{18.4} & 26.8 & 31.7 & \textbf{21.9} & 19.8 & 18.9 & 24.4 & 18.4 & -- & \textbf{27.2} & \textbf{19.2} \\
\midrule
\multirow{3}{*}{te}
& Supervised & NA & 4.8 & 13.2 & 15.1 & 8.8 & 8.7 & 8.8 & 12.9 & 6.7 & 3.2 & -- & 9  \\
%& GPT-3 6.7B & 32 & & & & & & & & & & & & \\
& \modelname & 4 & {{19.1}} & {{26.7}} & \textbf{{38.8}} & \textbf{{23.8}} & \textbf{{22.3}} & \textbf{{14.9}} & \textbf{{26.7}} & {{20.7}} & {{18.5}} & -- & \textbf{{20.9}}  \\
& \modelname & 4 & \textbf{19.5} & \textbf{29.4} & 37.9 & 23.6 & 21.6 & 7.2 & 25.8 & \textbf{21.3} & \textbf{25.2} & -- &20.5 \\
\midrule
\multirow{3}{*}{es}
& Supervised & NA & 12.1 & 29.3 & 25.1 & 21.0 & 23.9 & 18.1 & 18.5 & 7.1 & 0.4 & 0.5 & -- \\
%& GPT-3 6.7B & 32 & & & & & & & & & & & & \\
& \modelname & 1 & {{20.8}} & \textbf{{33.4}} & {{34.6}} & {{25.8}} & {{26.7}} & {{22.3}} & {{24.3}} & {{17.8}} & {{21.2}} & {{23.7}} & -- \\
& \modelname & 4 & \textbf{21.9} & 31.8 & \textbf{35.3} & \textbf{26.2} & \textbf{27.0} & \textbf{23.2} & \textbf{24.8} & \textbf{17.9} & \textbf{21.6} & \textbf{23.8} & -- \\
\bottomrule
\end{tabular}
}
\end{small}
\caption{Machine translation results on FLORES-101 devtest (spBLEU) using \modelname with 4-shots. Source language in rows, target language in columns. Supervised results correspond to the M2M-124 615M model from \citet{Goyal2022TheFE}. Bold denotes the best. spBLEU computed using the % official 
implementation from \citet{Goyal2022TheFE}.}
\label{tab:flores101_4shot}
\end{table*}

%% file: prompt_examples.tex
As described in Section~\ref{sec:eval}, we can use \modelname in different setting for in-context evaluation. In this section, we provide the details on the mode and the prompt used for all the tasks presented in this work. Tables~\ref{tab:prompt_examples1},  \ref{tab:prompt_examples2}, \ref{tab:prompt_examples3}, \ref{tab:prompt_winogender}, and \ref{tab:prompt_reasoning} present the evaluation mode and a sample prompt used for model evaluation. As we mentioned in Section~\ref{sec:english_nlp}, in zero-shot evaluation of \modelname, we realized that adding dummy examples can help the model generate output in the desired format related to each task. These dummy examples are denoted in red in the Table~\ref{tab:prompt_examples3}.

\begin{table*}[h!]
\centering
\begin{small}
\addtolength{\tabcolsep}{-2.5pt}
\scalebox{0.8}{
\begin{tabular}{lccl}
\toprule
Task & Shots & \makecell[c]{Evaluation \\ Mode} & Sample Prompt \\
\midrule
XSUM & 1 & \makecell[c]{CLM \\ (Generation)} & \makecell[l]{\ttfamily [CLM] Article: [dev set article]  ==> Short summary: [dev set summary] \\ \ttfamily <br><br><br> Article: [test set article] ==> Short summary:} \\\\
\makecell[l]{MLSUM \\ (de, es, fr)} & 1 & \makecell[c]{CLM \\ (Generation)} & \makecell[l]{\ttfamily [CLM] [dev set article]  ==> Summary: [dev set summary] <br><br><br> \\ \ttfamily [test set article] ==> Summary:} \\\\
E2E & 1 & \makecell[c]{CLM \\ (Generation)} & \makecell[l]{\ttfamily [CLM] name[Alimentum], area[city centre], familyFriendly[no] ==> sentence desribing \\ \ttfamily  the place: There is a place in the city centre, Alimentum, that is not family-friendly. ; \\ \ttfamily name[Blue Spice], eatType[coffee shop], area[city centre] ==> \\ \ttfamily sentence desribing the place:} \\\\
WebNLG & 1 & \makecell[c]{CLM \\ (Generation)} & \makecell[l]{\ttfamily [CLM] ['Aleksandra\_Kovac | genre | Soul\_music', 'Aleksandra\_Kovac | activeYearsStartYear | 1990'] \\ \ttfamily ==> triplets into natural language: Aleksandra Kovac stared in 1990 and she performs \\ 
\ttfamily soul music. <br> ['Nie\_Haisheng | birthDate | 1964-10-13', \\ \ttfamily 'Nie\_Haisheng | occupation | Fighter\_pilot'] ==> triplets into natural language:} \\\\
\\\\
\makecell[l]{Flores-101$^{**}$ \\ WMT'16 \\WMT'14 \\ WMT'19} & 1 & \makecell[c]{CLM \\ (Generation)} & \makecell[l]{\ttfamily [CLM] Sentence:  Elections européennes: Objectifs, financement...; \\ \ttfamily Translation in German: Europawahlen: Ziele, Finanzierung...; \\ \ttfamily Sentence:  Kipping au congrès de die Linke sur l'Europe : l'Europe est depuis longtemps \\ \ttfamily un continent d'immigration.; Translation in German:}\\\\
\bottomrule
\end{tabular}
}
\end{small}
\caption{Prompts used for \modelname 1-shot evaluation on different generation tasks. No newline characters are used in the prompts since our tokenizer removes that (newline breaks are used for display only). $^{**}$ For MT, we change the prompt for each language pair to reflect the correct target language.}
\label{tab:prompt_examples1}
\end{table*}

\begin{table*}[h!]
\centering
\begin{small}
\addtolength{\tabcolsep}{-2.5pt}
\scalebox{0.8}{
\begin{tabular}{lcl}
\toprule
Task & \makecell[c]{Evaluation \\ Mode} & Sample Prompt \\
\midrule
XNLI & \makecell[c]{Denoising \\ (Scoring)}  & 
\makecell[l]{
Encoder input: \\
\ttfamily The exhibition only displays cars from the 2000s, right? \\
\ttfamily It displays all kinds of vehicles, from the coach that carried Napoleon to and from Moscow in 1812  \\ 
\ttfamily to a splendid 4-horsepower Renault car from 1904 and other turn-of-the-century classics.\\ 
Decoder input: \\
\ttfamily The exhibition only displays cars from the 2000s, right? \textit{\{Yes, Also, No\}} \\
\ttfamily It displays all kinds of vehicles, from the coach that carried Napoleon to and from Moscow in 1812  \\ 
\ttfamily to a splendid 4-horsepower Renault car from 1904 and other turn-of-the-century classics.\\ 
}
\\\\
XCOPA & \makecell[c]{CLM \\ (Scoring)} & 
\makecell[l]{
Encoder input: \\
\ttfamily \textit{\{Ricevette la sua pensione, Ripagò il suo mutuo\}} perché \\
Decoder input: \\
\ttfamily La donna andò in pensione. \\
}
\\\\
Paws-X & \makecell[c]{Denoising \\ (Scoring)} & 
\makecell[l]{
Encoder input: \\
\ttfamily Joe R. Campa Jr. is a former sailor of the United States Navy, who served as the eleventh Master Chief \\
\ttfamily Petty Officer of the U.S. Navy, right? \\
\ttfamily Joe R. Campa Jr. is a former U.S. Navy Matrose who served as the 11th Master Chief Petty Officer \\
\ttfamily of the United States Navy.  \\ 
Decoder input: \\
\ttfamily Joe R. Campa Jr. is a former sailor of the United States Navy, who served as the eleventh Master Chief \\
\ttfamily Petty Officer of the U.S. Navy, right? \textit{\{Yes, No\}}\\
\ttfamily Joe R. Campa Jr. is a former U.S. Navy Matrose who served as the 11th Master Chief Petty Officer \\
\ttfamily of the United States Navy.  \\ 
}
\\\\
XWinograd & \makecell[c]{Denoising \\ (Scoring)} & 
\makecell[l]{
Encoder input: \\
\ttfamily In the storm, the tree fell down and crashed through the roof of my house. \\
\ttfamily Now, I have to get removed. \\
Decoder input: \\
\ttfamily In the storm, the tree fell down and crashed through the roof of my house. \\
\ttfamily Now, I have to get \textit{\{the tree, the roof\}} removed. \\ 
}\\
\bottomrule
\end{tabular}
}
\end{small}
\caption{Prompts used for \modelname zero-shot evaluation on multilingual datasets. No newline characters are used in the prompts since our tokenizer removes that (newline breaks are used for display only). Text in \textit{italics} indicate the options that are being scored. Except for XCOPA, we use a machine translation system to translate prompts from english to the target language. Evaluation mode is determined by performance on the validation set of each dataset (except Winograd, where we only use denoising).}
\label{tab:prompt_examples2}
\end{table*}

\begin{table*}[h!]
\centering
\begin{small}
\addtolength{\tabcolsep}{-2.5pt}
\scalebox{0.8}{
\begin{tabular}{lccl}
\toprule
Task & Shots & \makecell[c]{Evaluation \\ Mode} & Sample Prompt \\
\midrule
SQuADv2 & 0 & \makecell[c]{CLM \\ (Generation)} & \makecell[l]{\ttfamily [CLM] Context: The Normans (Norman: Nourmands; French: Normands; Latin: Normanni) \\
\ttfamily were the people who in the 10th and 11th centuries gave their name to Normandy, a region in France. \\ \ttfamily They were ... succeeding centuries.\\ \color{red}\ttfamily Question: What is the last word in the passage? Answer: centuries; \\ \ttfamily Question: In what country is Normandy located? Answer:}\\\\
\\\\
BoolQ & 0 & \makecell[c]{CLM \\ (Generation)} & \makecell[l]{ \ttfamily [CLM] Context: All biomass goes through at least ... producing gasoline. <br>\\ \color{red} \ttfamily Question: Is this passage written in English? <br>  Answer (True or False): True <br>\\ \color{red} \ttfamily Question: Is this passage written in French? <br>  Answer (True or False): False <br>\\ \ttfamily Question: does ethanol take more energy make that produces <br> Answer (True or False):}\\\\
CB & 0 & \makecell[c]{CLM \\ (Generation)} & \makecell[l]{ \ttfamily [CLM] Context: It grew bigger with incredible speed ... a child a toddler with a red woolly hat on. <br>\\ \color{red} \ttfamily Question: Is this passage written in English? <br>  Answer (True, False, or Neither): True <br>\\ \color{red} \ttfamily Question: Is this passage written in French? <br>  Answer (True, False, or Neither): False <br>\\ \ttfamily Question: it was a child <br> Answer (True, False, or Neither):} \\\\
RTE & 0 & \makecell[c]{CLM \\ (Generation)} &  \makecell[l]{\ttfamily [CLM] Context: Dana Reeve, the widow of the actor Christopher Reeve, has died of lung cancer at age 44, \\ \ttfamily according to the Christopher Reeve Foundation. <br> \\\ttfamily \color{red} The first word in the context is "Dana". True or False? True <br> \\ \ttfamily \color{red} The first word in the context is "Reeve". True or False? False <br> \\ \ttfamily Christopher Reeve had an accident. True or False?}\\\\
ReCoRD & 0 & \makecell[c]{Denoising \\ (Scoring)} & 
\makecell[l]{
Encoder input: \\
\ttfamily Tracy Morgan hasn't appeared on stage since the devastating New Jersey crash ... \\
\ttfamily new SNL season will be hosted by Miley Cyrus, followed by Amy Schumer <br> \\
\ttfamily On October 10, acclaimed comedian and star of the summer box office hit Trainwreck \\
\ttfamily Amy Schumer will make her SNL debut, followed by a week later. \\
Decoder input: \\
\ttfamily Tracy Morgan hasn't appeared on stage since the devastating New Jersey crash ... \\
\ttfamily new SNL season will be hosted by Miley Cyrus, followed by Amy Schumer <br> \\
\ttfamily On October 10, acclaimed comedian and star of the summer box office hit Trainwreck \\
\ttfamily Amy Schumer will make her SNL debut, followed by \textit{\{Amy Schumer, James, Jimmy Mack,} \\
\ttfamily \textit{McNair, Miley Cyrus, Morgan, NBC, New Jersey, New Jersey Turnpike, Night Live, SNL,} \\
\ttfamily \textit{Season 41, Tracy Morgan, Twitter\}} a week later. \\
}\\\\
WSC & 0 & \makecell[c]{Denoising \\ (Scoring)} & \makecell[l]{\ttfamily Passage: Bernard , who had not told the government official that he was less than 21 when he \\\ttfamily filed for a homestead claim,  did not consider that he had  ... his claim away from *him* . <br> \\\ttfamily \color{red} Question: The first word in the context is "Bernard". True or False ? Answer: True <br> \\ \ttfamily \color{red} Question: The first word in the context is "19". True or False ? Answer: False <br> \\ \ttfamily Question: The pronoun "*him*" in the passage refers to anyone. True or False ? Answer:} \\\\
WiC & 0 & \makecell[c]{CLM \\ (Generation)} & \makecell[l]{\ttfamily [CLM] \color{red} Sentence 1: life is beautiful. Sentence 2: sky is blue. ; \\ \ttfamily \color{red} Question: Are Sentence 1 and Sentence 2   in the same language? Answer ( yes or no ): yes <br> \\ \ttfamily Sentence 1: An emerging professional class. Sentence 2: Apologizing for losing your temper, \\ \ttfamily even though you were badly provoked, showed real class. ; \\ \ttfamily Question: Does the word " class " in Sentence 1 have the same meaning as in Sentence 2 ? \\ \ttfamily Answer ( yes or no ):}\\\\
CoPA & 0 & \makecell[c]{CLM \\ (Scoring)} & \makecell[l]{Encoder input: \\ \ttfamily [CLM] (Choice 1/2) \textit{because}/\textit{so} \\
Decoder Input: \\
\ttfamily (Premise) %\\ \ttfamily (Choice 2) \textit{because}/\textit{so} (Premise)}
}
\\\\
MultiRC & 0 &  \makecell[c]{CLM \\ (Generation)} & \makecell[l]{\ttfamily [CLM] Context: What causes a change in motion? ... This requires only a very small force. <br> \\
\ttfamily Based on the Context determine whether following Question and Answer pairs are correct or incorrect: \\
\ttfamily \color{red} Q: What language is the context written in? A: English (correct) ;  \\
\ttfamily \color{red} Q: What is the first word in the context? A: small (incorrect) ;  \\
\ttfamily Q: Would the mass of a baseball affect how much force you have to use to pick it up? A: No}\\
\bottomrule
\end{tabular}
}
\end{small}
\caption{Prompts used for \modelname zero-shot evaluation on English datasets. No newline characters are used in the prompts since our tokenizer removes that (newline breaks are used for display only). The dummy examples are denoted by red text. Ellipses are for brevity only in this table; the full text was used in the actual prompts to the model. }
\label{tab:prompt_examples3}
\end{table*}

\begin{table*}[h!]
\centering
\begin{small}
\addtolength{\tabcolsep}{-2.5pt}
\scalebox{0.8}{
\begin{tabular}{lccl}
\toprule
Task & Shots & \makecell[c]{Evaluation \\ Mode} & Sample Prompt \\
\midrule
Winogender & 0 & \makecell[c]{Denoising \\ (Scoring)} & 
\makecell[l]{
Encoder input: \\
\ttfamily Context: The customer ordered a chocolate cake from the baker even though she had bad reviews \\
\ttfamily online. Question: "she" refers to? Answer:\\
Decoder input: \\
\ttfamily Context: The customer ordered a chocolate cake from the baker even though she had bad reviews \\
\ttfamily online. Question: "she" refers to? Answer: baker
}\\

Winogender & 0 & \makecell[c]{CLM \\ (Scoring)} & 
\makecell[l]{
Encoder input: \\
\ttfamily [CLM] Context: The customer ordered a chocolate cake from the baker even though she had bad reviews \\
\ttfamily online. Question: "she" refers to? Answer:\\
Decoder input: \\
\ttfamily baker
}\\
\bottomrule
\end{tabular}
}
\end{small}
\caption{Prompts used for \modelname zero-shot evaluation on Winogender. No newline characters are used in the prompts since our tokenizer removes that (newline breaks are used for display only).}
\label{tab:prompt_winogender}
\end{table*}

\begin{table*}[h!]
\centering
\begin{small}
\addtolength{\tabcolsep}{-2.5pt}
\scalebox{0.8}{
\begin{tabular}{lcl}
\toprule
Task & \makecell[c]{Evaluation \\ Mode} & Sample Prompt \\
\midrule
MultiArith & \makecell[c]{Zero-shot} & 
\makecell[l]{
Encoder input:
\ttfamily [CLM] Question: If there are 3 cars in the parking lot and 2 more cars arrive, how many 
\\ \ttfamily cars are in the parking lot? <br> Answer: \\
}\\
\midrule
MultiArith & \makecell[c]{Zero-shot \\ chain-of-thought} & 
\makecell[l]{
Encoder input (round-1): 
\ttfamily [CLM] Question: If there are 3 cars in the parking lot and 2 more cars arrive, how many 
\\ \ttfamily cars are in the parking lot? <br> Show your work. \\
Encoder input (round-2): 
\ttfamily [CLM] Question: If there are 3 cars in the parking lot and 2 more cars arrive, how many 
\\ \ttfamily cars are in the parking lot? 
\\ \ttfamily <br> Show your work. <br> [round-1-output] <br> Therefore, the answer is
}\\
\bottomrule
\end{tabular}
}
\end{small}
\caption{Prompts used for \modelname zero-shot chain-of-thought evaluation on MultiArith. No newline characters are used in the prompts since our tokenizer removes that (newline breaks are used for display only).}
\label{tab:prompt_reasoning}
\end{table*}

%% file: filtered_datasets.tex
Below is the list of datasets filtered from the \modelname training data with the names as appear in Hugging Face's datasets hub~\citep{lhoest-etal-2021-datasets}. If a dataset is available in different versions or languages, all are filtered.
\begin{ttfamily}
\begin{itemize}
\item amazon\_reviews\_multi
\item assin2
\item cbt
\item coqa
\item drop
\item glue
\item hellaswag
\item lambada
\item mkb
\item mlqa
\item paws-x
\item race
\item squad
\item super\_glue
\item wiki\_atomic\_edits
\item xnli
\item xquad
 \end{itemize}
 \end{ttfamily}